\documentclass[runningheads]{llncs}

\newif\ifarxiv
\arxivtrue

\newif\iffinal
\finaltrue

\newif\ifextended
\extendedfalse

\usepackage{graphicx}
\usepackage[hyphens]{url}
\usepackage[colorlinks=true,allcolors=blue]{hyperref}

\usepackage[caption=false,farskip=0pt]{subfig}
\usepackage{booktabs}
\usepackage{diagbox}
\usepackage{multirow}
\usepackage[acronym]{glossaries}
\usepackage{arydshln} %
\usepackage{xspace}
\usepackage[utf8]{inputenc}
\usepackage[T1]{fontenc}
\usepackage[capitalise]{cleveref} %
\usepackage[dvipsnames,table]{xcolor}

\usepackage{cite} %
\usepackage{ulem} %
\usepackage{enumitem}

\makeatletter
\newcommand*\mysize{%
  \@setfontsize\mysize{8.5}{8.5}%
}
\makeatother

\newacronymstyle{long-short-br}
{%
  \GlsUseAcrEntryDispStyle{long-short}%
}%
{%
  \GlsUseAcrStyleDefs{long-short}%
}
\setacronymstyle{long-short-br}

\newcommand*{\RL}[2][]{\textcolor{Rhodamine}{[\textbf{\ifthenelse{\equal{#1}{}}{RL}{RL(#1)}}: #2]}}
\newcommand*{\DM}[2][]{\textcolor{blue}{[\textbf{\ifthenelse{\equal{#1}{}}{DM}{DM(#1)}}: #2]}}
\newcommand*{\LZ}[2][]{\textcolor{Green}{[\textbf{\ifthenelse{\equal{#1}{}}{LZ}{LZ(#1)}}: #2]}}
\newcommand*{\RM}[2][]{\textcolor{purple}{[\textbf{\ifthenelse{\equal{#1}{}}{RM}{RM(#1)}}: #2]}}
\newcommand*{\todo}[2][]{\textcolor{red}{[\textbf{\ifthenelse{\equal{#1}{}}{TODO}{TODO(#1)}}: #2]}}

\newcommand\red[1]{{\textcolor{red}{#1}}}

\newcommand\customEmph[1]{{\textit{#1}}}

\newcommand\topcross[1]{{#1}} %

\newlength{\adj}
\usepackage{transparent}
\usepackage{tikz}
\ifarxiv
\newcommand\copyrighttext{%
  \scriptsize Accepted at CIARP 2023. This is an author-prepared version. The final publication is available on \textit{Springer} (DOI: \href{https://doi.org/10.1007/978-3-031-49249-5_5}{\textcolor{blue}{10.1007/978-3-031-49249-5\_5}}).}
\newcommand\copyrightnotice{%
\vspace{-5mm}
\begin{tikzpicture}[remember picture,overlay]
\node[anchor=south,yshift=80pt,xshift=0pt] at (current page.south) {\fbox{\transparent{0.85}\parbox{\dimexpr0.84\textwidth-\fboxsep-\fboxrule\relax}{\copyrighttext}}};
\end{tikzpicture}%
}
\else
\fi

\begin{document}

\newacronym{alpr}{ALPR}{Automatic License Plate Recognition}
\newacronym{bilstm}{Bi-LSTM}{Bi-directional Long Short-Term Memory}
\newacronym{cnn}{CNN}{Convolutional Neural Network}
\newacronym{crnn}{CRNN}{Convolutional Recurrent Neural Network}
\newacronym{ctc}{CTC}{Connectionist Temporal Classification}
\newacronym{denatran}{DENATRAN}{National Traffic Department of Brazil}
\newacronym{fps}{FPS}{frames per second}
\newacronym{gan}{GAN}{Generative Adversarial Network}
\newacronym{grcnn}{GRCNN}{Gated Recurrent Convolution Neural Network}
\newacronym{iou}{IoU}{Intersection over Union}
\newacronym{it}{IT}{information technology}
\newacronym{lodo}{LODO}{leave-one-dataset-out}
\newacronym{lpd}{LPD}{license plate detection}
\newacronym{lpr}{LPR}{license plate recognition}
\newacronym{lprAbstract}{LPR}{License Plate Recognition}
\newacronym{lp}{LP}{license plate}
\newacronym{lstm}{LSTM}{Long Short-Term Memory}
\newacronym{nlp}{NLP}{Natural Language Processing}
\newacronym{ocr}{OCR}{Optical Character Recognition}
\newacronym{rare}{RARE}{Robust text recognizer with Automatic REctification}
\newacronym{rcnn}{RCNN}{Recurrent Convolutional Neural Network}
\newacronym{resnet}{ResNet}{Residual Network}
\newacronym{rodosol}{RodoSol}{\textit{Rodovia do Sol}}
\newacronym{rtwoam}{R\textsuperscript{2}AM}{Recursive Recurrent neural networks with Attention Modeling}
\newacronym{sgd}{SGD}{Stochastic Gradient Descent}
\newacronym{starnet}{STAR-Net}{SpaTial Attention Residue Network}
\newacronym{stn}{STN}{Spatial Transformer Network}
\newacronym{tps}{TPS}{thin-plate splines}
\newacronym{trba}{TRBA}{TPS-ResNet-BiLSTM-Attention}
\newacronym{vit}{ViT}{Vision Transformer}

\newacronym{bm}{BM}{Best Model}
\newacronym{hc}{HC}{Highest Confidence}
\newacronym{mv}{MV}{Majority Vote}
\newacronym{mvcp}{MVCP}{Majority Vote by Character Position}
\newcommand{\hc}{HC\xspace}
\newcommand{\mvbm}{MV--BM\xspace}
\newcommand{\mvhc}{MV--HC\xspace}
\newcommand{\mvcp}{MVCP\xspace}
\newcommand{\mvcpbm}{MVCP--BM\xspace}
\newcommand{\mvcphc}{MVCP--HC\xspace}

\newcommand{\aolp}{AOLP\xspace}
\newcommand{\caltech}{Caltech Cars\xspace}
\newcommand{\ccpd}{CCPD\xspace}
\newcommand{\cdhard}{CD-HARD\xspace}
\newcommand{\chineselp}{ChineseLP\xspace}
\newcommand{\clpd}{CLPD\xspace}
\newcommand{\englishlp}{EnglishLP\xspace}
\newcommand{\openalprbr}{OpenALPR-BR\xspace}
\newcommand{\openalpreu}{OpenALPR-EU\xspace}
\newcommand{\pku}{PKU\xspace}
\newcommand{\rodosol}{\rodosolalpr}
\newcommand{\rodosolalpr}{RodoSol-ALPR\xspace}
\newcommand{\ssigsegplate}{SSIG-SegPlate\xspace}
\newcommand{\stills}{UCSD-Stills\xspace}
\newcommand{\ufop}{UFOP\xspace}
\newcommand{\ufpralpr}{UFPR-ALPR\xspace}

\newcommand{\model}{DC-NET\xspace}

\newcommand{\crnn}{\acrshort*{crnn}\xspace}
\newcommand{\grcnn}{\acrshort*{grcnn}\xspace}
\newcommand{\rare}{\acrshort*{rare}\xspace}
\newcommand{\rosetta}{Rosetta\xspace}
\newcommand{\rtwoam}{\acrshort*{rtwoam}\xspace}
\newcommand{\starnet}{\acrshort*{starnet}\xspace}
\newcommand{\trba}{\acrshort*{trba}\xspace}
\newcommand{\vitstrbase}{ViTSTR-Base\xspace}
\newcommand{\vitstrsmall}{ViTSTR-Small\xspace}
\newcommand{\vitstrtiny}{ViTSTR-Tiny\xspace}
\newcommand{\vitstr}{ViTSTR\xspace}

\newcommand{\holistic}{Holistic-CNN\xspace}
\newcommand{\multitaskgabriel}{Multi-Task-LR\xspace}

\newcommand{\crnet}{CR-NET\xspace}
\newcommand{\fastocr}{Fast-OCR\xspace}
\newcommand{\nummodels}{12\xspace}
\newcommand{\numbaselines}{\nummodels}
\newcommand{\numdatasets}{12\xspace}
\newcommand{\numdatasetsTrad}{8\xspace}
\newcommand{\numdatasetsTradWords}{eight\xspace}
\newcommand{\numdatasetsCross}{4\xspace}
\newcommand{\numdatasetsCrossWords}{four\xspace}

\newcommand{\bestIndividual}{$92.4$\%\xspace}
\newcommand{\errorIndividual}{$7.6$\%\xspace}
\newcommand{\bestFusion}{$97.6$\%\xspace}
\newcommand{\errorFusion}{$2.4$\%\xspace}
\newcommand{\errorReduction}{$68.4$\%\xspace}

\newcommand{\bestIndividualCross}{$87.6$\%\xspace}
\newcommand{\bestFusionCross}{$90.3$\%\xspace}
\newcommand{\fusionExceeding}{$90$\%\xspace}

\iffinal
\newcommand{\urlSupplementary}{\url{https://raysonlaroca.github.io/supp/lpr-model-fusion/}}
\else
\newcommand{\urlSupplementary}{\url{https://github.com/anonymous_author/placeholder/}}
\fi

\newcommand{\accAvgDefaultCaltech}{92.0\xspace}
\newcommand{\accAvgDefaultEnglishlp}{88.0\xspace}
\newcommand{\accAvgDefaultStills}{92.2\xspace}
\newcommand{\accAvgDefaultChineselp}{94.3\xspace}
\newcommand{\accAvgDefaultAolp}{97.4\xspace}
\newcommand{\accAvgDefaultSsig}{95.0\xspace}
\newcommand{\accAvgDefaultUfpr}{77.7\xspace}
\newcommand{\accAvgDefaultRodosol}{79.8\xspace}
\newcommand{\accAvgDefaultTrad}{89.6\xspace}

\newcommand{\accAvgHcbmCaltech}{97.8\xspace}
\newcommand{\accAvgHcbmEnglishlp}{95.1\xspace}
\newcommand{\accAvgHcbmStills}{96.7\xspace}
\newcommand{\accAvgHcbmChineselp}{98.1\xspace}
\newcommand{\accAvgHcbmAolp}{99.0\xspace}
\newcommand{\accAvgHcbmSsig}{96.6\xspace}
\newcommand{\accAvgHcbmUfpr}{90.9\xspace}
\newcommand{\accAvgHcbmRodosol}{93.5\xspace}
\newcommand{\accAvgHcbmTrad}{96.0\xspace}

\newcommand{\accAvgHcmvCaltech}{97.8\xspace}
\newcommand{\accAvgHcmvEnglishlp}{95.1\xspace}
\newcommand{\accAvgHcmvStills}{96.7\xspace}
\newcommand{\accAvgHcmvChineselp}{98.1\xspace}
\newcommand{\accAvgHcmvAolp}{99.0\xspace}
\newcommand{\accAvgHcmvSsig}{96.6\xspace}
\newcommand{\accAvgHcmvUfpr}{90.9\xspace}
\newcommand{\accAvgHcmvRodosol}{93.5\xspace}
\newcommand{\accAvgHcmvTrad}{96.0\xspace}

\newcommand{\accAvgMvbmCaltech}{97.8\xspace}
\newcommand{\accAvgMvbmEnglishlp}{97.1\xspace}
\newcommand{\accAvgMvbmStills}{100.0\xspace}
\newcommand{\accAvgMvbmChineselp}{98.1\xspace}
\newcommand{\accAvgMvbmAolp}{99.7\xspace}
\newcommand{\accAvgMvbmSsig}{98.4\xspace}
\newcommand{\accAvgMvbmUfpr}{92.7\xspace}
\newcommand{\accAvgMvbmRodosol}{96.4\xspace}
\newcommand{\accAvgMvbmTrad}{97.5\xspace}

\newcommand{\accAvgMvhcCaltech}{97.8\xspace}
\newcommand{\accAvgMvhcEnglishlp}{97.1\xspace}
\newcommand{\accAvgMvhcStills}{100.0\xspace}
\newcommand{\accAvgMvhcChineselp}{98.1\xspace}
\newcommand{\accAvgMvhcAolp}{99.7\xspace}
\newcommand{\accAvgMvhcSsig}{99.1\xspace}
\newcommand{\accAvgMvhcUfpr}{92.3\xspace}
\newcommand{\accAvgMvhcRodosol}{96.5\xspace}
\newcommand{\accAvgMvhcTrad}{97.6\xspace}

\newcommand{\accAvgMvhcAltCaltech}{97.8\xspace}
\newcommand{\accAvgMvhcAltEnglishlp}{97.1\xspace}
\newcommand{\accAvgMvhcAltStills}{98.3\xspace}
\newcommand{\accAvgMvhcAltChineselp}{98.1\xspace}
\newcommand{\accAvgMvhcAltAolp}{99.6\xspace}
\newcommand{\accAvgMvhcAltSsig}{99.4\xspace}
\newcommand{\accAvgMvhcAltUfpr}{91.8\xspace}
\newcommand{\accAvgMvhcAltRodosol}{96.2\xspace}
\newcommand{\accAvgMvhcAltTrad}{97.3\xspace}

\newcommand{\accAvgMvcpBmCaltech}{95.7\xspace}
\newcommand{\accAvgMvcpBmEnglishlp}{96.1\xspace}
\newcommand{\accAvgMvcpBmStills}{100.0\xspace}
\newcommand{\accAvgMvcpBmChineselp}{98.1\xspace}
\newcommand{\accAvgMvcpBmAolp}{99.6\xspace}
\newcommand{\accAvgMvcpBmSsig}{99.0\xspace}
\newcommand{\accAvgMvcpBmUfpr}{92.8\xspace}
\newcommand{\accAvgMvcpBmRodosol}{96.4\xspace}
\newcommand{\accAvgMvcpBmTrad}{97.2\xspace}

\newcommand{\accAvgMvcpHcCaltech}{97.8\xspace}
\newcommand{\accAvgMvcpHcEnglishlp}{96.1\xspace}
\newcommand{\accAvgMvcpHcStills}{100.0\xspace}
\newcommand{\accAvgMvcpHcChineselp}{98.1\xspace}
\newcommand{\accAvgMvcpHcAolp}{99.6\xspace}
\newcommand{\accAvgMvcpHcSsig}{99.3\xspace}
\newcommand{\accAvgMvcpHcUfpr}{92.5\xspace}
\newcommand{\accAvgMvcpHcRodosol}{96.3\xspace}
\newcommand{\accAvgMvcpHcTrad}{97.5\xspace}

\newcommand{\accAvgDefaultOpenalprBR}{97.3\xspace}
\newcommand{\accAvgDefaultOpenalprEU}{92.0\xspace}
\newcommand{\accAvgDefaultPku}{98.3\xspace}
\newcommand{\accAvgDefaultCdhard}{32.4\xspace}
\newcommand{\accAvgDefaultClpd}{90.6\xspace}
\newcommand{\accAvgDefaultCross}{78.1\xspace}

\newcommand{\accAvgHcbmCrossOpenalprEU}{95.4\xspace}
\newcommand{\accAvgHcbmCrossPku}{99.2\xspace}
\newcommand{\accAvgHcbmCrossCdhard}{48.1\xspace}
\newcommand{\accAvgHcbmCrossClpd}{94.9\xspace}
\newcommand{\accAvgHcbmCrossCross}{84.4\xspace}

\newcommand{\accAvgHcmvCrossOpenalprBR}{98.2\xspace}
\newcommand{\accAvgHcmvCrossOpenalprEU}{95.4\xspace}
\newcommand{\accAvgHcmvCrossPku}{99.2\xspace}
\newcommand{\accAvgHcmvCrossCdhard}{48.1\xspace}
\newcommand{\accAvgHcmvCrossClpd}{94.9\xspace}
\newcommand{\accAvgHcmvCrossCross}{84.4\xspace}

\newcommand{\accAvgMvbmCrossOpenalprBR}{98.2\xspace}
\newcommand{\accAvgMvbmCrossOpenalprEU}{99.1\xspace}
\newcommand{\accAvgMvbmCrossPku}{99.7\xspace}
\newcommand{\accAvgMvbmCrossCdhard}{65.4\xspace}
\newcommand{\accAvgMvbmCrossClpd}{97.0\xspace}
\newcommand{\accAvgMvbmCrossCross}{90.3\xspace}

\newcommand{\accAvgMvcpBmCrossOpenalprEU}{95.4\xspace}
\newcommand{\accAvgMvcpBmCrossPku}{99.7\xspace}
\newcommand{\accAvgMvcpBmCrossCdhard}{54.8\xspace}
\newcommand{\accAvgMvcpBmCrossClpd}{95.5\xspace}
\newcommand{\accAvgMvcpBmCrossCross}{86.3\xspace}

\newcommand{\accAvgMvcpHcCrossOpenalprEU}{97.2\xspace}
\newcommand{\accAvgMvcpHcCrossPku}{99.7\xspace}
\newcommand{\accAvgMvcpHcCrossCdhard}{57.7\xspace}
\newcommand{\accAvgMvcpHcCrossClpd}{95.9\xspace}
\newcommand{\accAvgMvcpHcCrossCross}{87.6\xspace}

\newcommand{\accAvgMvbmAltCrossOpenalprBR}{98.2\xspace}
\newcommand{\accAvgMvbmAltCrossOpenalprEU}{99.1\xspace}
\newcommand{\accAvgMvbmAltCrossPku}{99.7\xspace}
\newcommand{\accAvgMvbmAltCrossCdhard}{68.3\xspace}
\newcommand{\accAvgMvbmAltCrossClpd}{96.3\xspace}
\newcommand{\accAvgMvbmAltCrossCross}{90.8\xspace}

\newcommand{\accAvgMvhcCrossOpenalprBR}{98.2\xspace}
\newcommand{\accAvgMvhcCrossOpenalprEU}{99.1\xspace}
\newcommand{\accAvgMvhcCrossPku}{99.7\xspace}
\newcommand{\accAvgMvhcCrossCdhard}{65.4\xspace}
\newcommand{\accAvgMvhcCrossClpd}{96.3\xspace}
\newcommand{\accAvgMvhcCrossCross}{90.1\xspace}

\newcommand{\paperTitle}{Leveraging Model Fusion for\\Improved License Plate Recognition}

\newcommand\customorcidAuthor[1]{\hspace*{-1mm}
\href{https://orcid.org/#1}{\includegraphics[width=0.3cm]{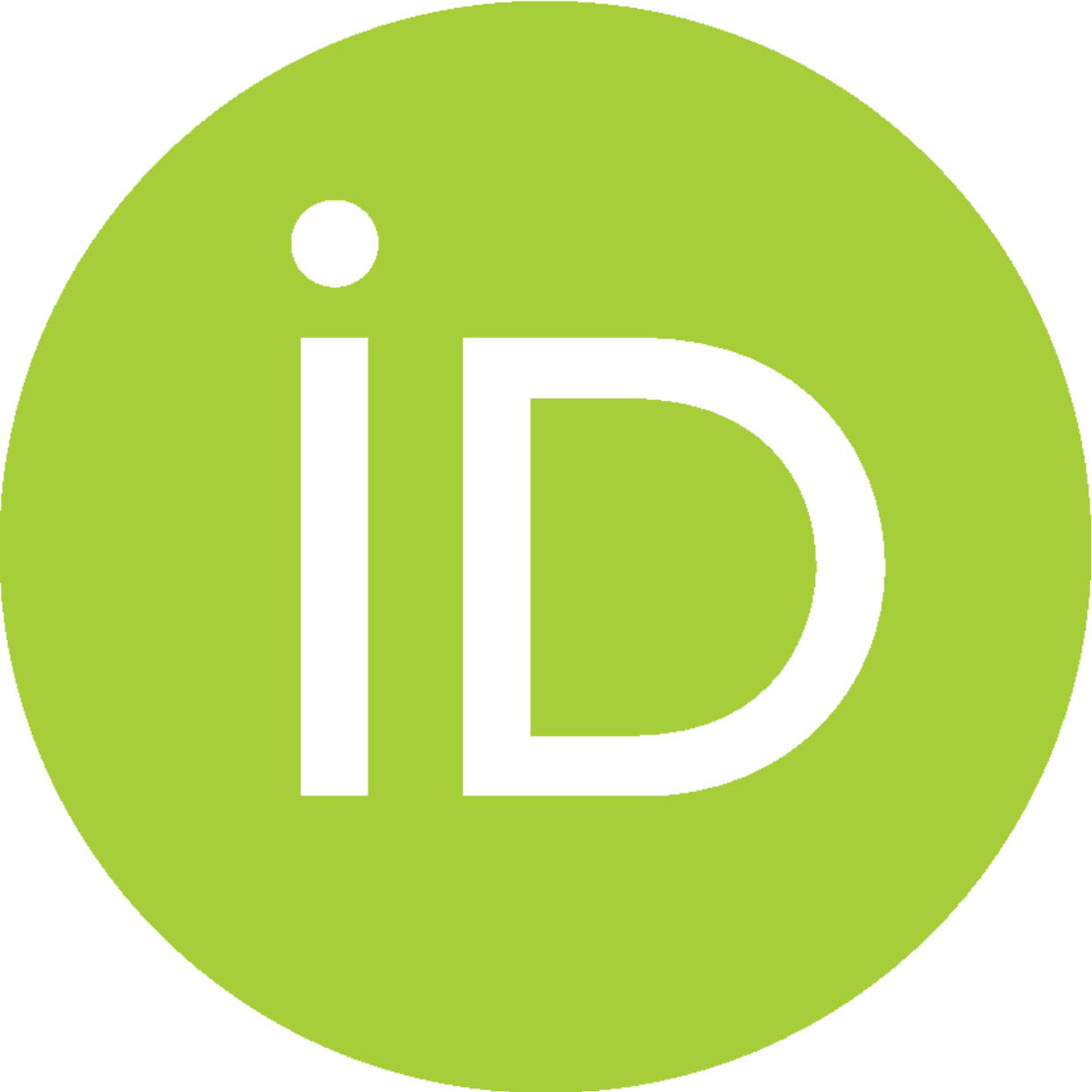}}\hspace*{-1mm}
}

\iffinal
\title{\paperTitle\thanks{Supported by the Coordination for the Improvement of Higher Education Personnel~(CAPES) (\#~88881.516265/2020-01), and by the National Council for Scientific and Technological Development~(CNPq) (\#~309953/2019-7 and \#~308879/2020-1).}}
\else
\title{\paperTitle}
\fi
\iffinal
\author{Rayson Laroca\inst{1}\customorcidAuthor{0000-0003-1943-2711}~\and Luiz A. Zanlorensi\inst{1}\customorcidAuthor{0000-0003-2545-0588}~\and Valter Estevam\inst{1,2}\customorcidAuthor{0000-0001-9491-5882}~\and Rodrigo~Minetto\inst{3}\customorcidAuthor{0000-0003-2277-4632}~\and David~Menotti\inst{1}\customorcidAuthor{0000-0003-2430-2030}
}

\authorrunning{R. Laroca et al.}

\institute{Federal University of Paran\'a, Curitiba, Brazil \and
Federal Institute of Paran\'a, Irati, Brazil \and
Federal University of Technology-Paran\'a, Curitiba, Brazil}
\else
\author{Anonymous Authors}

\authorrunning{Anonymous Authors}

\institute{Anonymous Affiliations}
\fi
\maketitle              %

\ifarxiv
\copyrightnotice
\else
\fi

\begin{abstract}
\gls*{lprAbstract} plays a critical role in various applications, such as toll collection, parking management, and traffic law enforcement.
Although \gls*{lprAbstract} has witnessed significant advancements through the development of deep learning, there has been a noticeable lack of studies exploring the potential improvements in results by fusing the outputs from multiple recognition models.
This research aims to fill this gap by investigating the combination of up to 12 different models using straightforward approaches, such as selecting the most confident prediction or employing majority vote-based strategies.
Our experiments encompass a wide range of datasets, revealing substantial benefits of fusion approaches in both intra- and cross-dataset setups.
Essentially, fusing multiple models reduces considerably the likelihood of obtaining subpar performance on a particular dataset/scenario.
We also found that combining models based on their speed is an appealing approach.
Specifically, for applications where the recognition task can tolerate some additional time, though not excessively, an effective strategy is to combine $4$--$6$ models.
These models may not be the most accurate individually, but their fusion strikes an optimal balance between speed and~accuracy.
\keywords{License Plate Recognition \and Model Fusion \and Ensemble.}
\end{abstract}

\section{Introduction}
\label{sec:introduction}

\glsresetall
\setcounter{footnote}{0}

\gls*{alpr} has garnered substantial interest in recent years due to its many practical applications, which include toll collection, parking management, border control, and road traffic monitoring~\cite{laroca2021efficient,wang2022rethinking,ke2023ultra}.

In the deep learning era, \gls*{alpr} systems customarily comprise two key components: \gls*{lpd} and \gls*{lpr}.
\gls*{lpd} entails locating regions containing \glspl*{lp} within an image, while \gls*{lpr} involves identifying the characters within these \glspl*{lp}.
Recent research has predominantly concentrated on advancing \gls*{lpr}~\cite{liu2021fast,zhang2021robust_attentional,nascimento2023super}, given that widely adopted object detectors such as Faster-RCNN and YOLO have consistently delivered impressive results in \gls*{lpd} for some years now~\cite{hsu2017robust,laroca2018robust,zhang2018joint}.

This study also focuses on \gls*{lpr} but provides a unique perspective compared to recent research.
Although deep learning techniques have enabled significant advancements in this field over the past years, multiple studies have shown that different models exhibit varying levels of robustness across different datasets~\cite{zeni2020weakly,mokayed2021new,laroca2022cross}.
Each dataset poses distinct challenges, such as diverse \gls*{lp} layouts and varying tilt ranges.
As a result, a method that performs optimally on one dataset may yield poor results on another.
This raises an important question: \customEmph{``Can we substantially enhance LPR results by fusing the outputs of diverse recognition models?''}
If so, two additional questions arise: \customEmph{``To what extent can this improvement be attained?''} and \customEmph{``How many and which models should be employed?''}
As of now, such questions remain unanswered in the existing~literature.

We acknowledge that some \gls*{alpr} applications impose stringent time constraints on their execution.
This is particularly true for embedded systems engaged in tasks such as access control and parking management in high-traffic areas.
However, in other contexts, such as systems used for issuing traffic tickets and conducting forensic investigations, there is often a preference to prioritize the recognition rate, even if it sacrifices efficiency~\cite{izidio2020embedded,nascimento2023super,schirrmacher2023benchmarking}.
These scenarios can greatly benefit from the fusion of multiple recognition~models.

Taking this into account, in this study, we thoroughly examine the potential of enhancing \gls*{lpr} results through the fusion of outputs from multiple recognition models. 
Remarkably, we assess the combination of up to \numbaselines well-known models across \numdatasets different datasets, setting our investigation apart from earlier~studies.

In summary, this paper has two main contributions:
\renewcommand{\labelitemi}{$\bullet$}
\begin{itemize}[leftmargin=3em]
    \item We present empirical evidence showcasing the benefits offered by fusion approaches in both intra- and cross-dataset setups.
    In the intra-dataset setup, the mean recognition rate across the datasets experiences a substantial boost, rising from~\bestIndividual achieved by the best model individually to~\bestFusion when leveraging the best fusion approach.
    Similarly, in the cross-dataset setup, the mean recognition rate increases from~\bestIndividualCross to levels exceeding~\fusionExceeding.
    Notably, in both setups, the sequence-level majority vote fusion approach outperform both character-level majority vote and selecting the prediction made with the highest~confidence approaches.
    \item We draw attention to the effectiveness of fusing models based on their speed.
    This approach is particularly useful for applications where the recognition task can accommodate a moderate increase in processing time.
    In such cases, the recommended strategy is to combine $4$--$6$ fast models.
    Although these models may not achieve the highest accuracy individually, their fusion results in an optimal trade-off between speed and~accuracy.
\end{itemize}

The rest of this paper is structured as follows.
\cref{sec:related_work} provides a concise overview of the recognition models explored in this work.
The experimental setup adopted in our research is detailed in \cref{sec:experimental-setup}.
The results obtained are presented and analyzed in \cref{sec:results}.
Lastly, \cref{sec:conclusions} summarizes our~findings.
\section{Related Work}
\label{sec:related_work}

\gls*{lpr} is widely recognized as a specific application within the field of scene text recognition~\cite{zou2020robust,lee2022license,gao2023group}.
\gls*{lpr} sets itself apart primarily due to the limited presence of strong linguistic context information and the minimal variation observed between characters.
The following paragraphs briefly describe well-known models originally proposed for general scene text recognition, \gls*{lpr}, and related tasks.
These models will be explored in this~study.

\begin{figure}[!htb]
    \centering
    
    \vspace{-3mm}
    
    \includegraphics[width=0.9\linewidth]{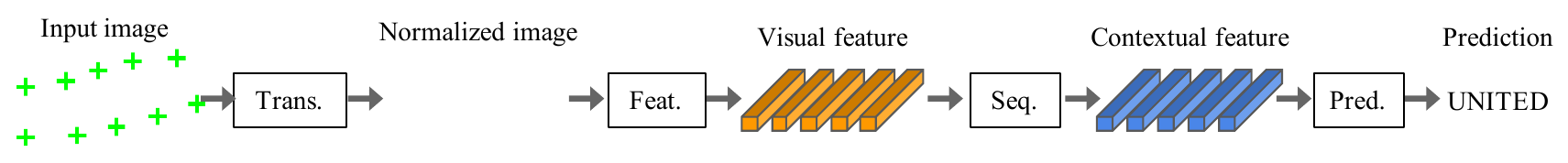}

    \vspace{-4.6mm}
    
    \caption{The four modules or stages of modern scene text recognition, according to~\cite{baek2019what}. ``Trans.'' stands for Transformation, ``Feat.'' stands for Feature Extraction, ``Seq.'' stands for Sequence Modeling, and ``Pred.'' stands for Prediction. Image reproduced from~\cite{baek2019what}.}
    \label{fig:related_work-miscellaneous:baek2019what:four-stages-str}

    \vspace{-1mm}
\end{figure}

Baek et al.~\cite{baek2019what} introduced a four-stage framework (depicted in \cref{fig:related_work-miscellaneous:baek2019what:four-stages-str}) that models the design patterns of most modern methods for scene text recognition.
The \customEmph{Transformation} stage removes the distortion from the input image so that the text is horizontal or normalized.
This task is generally done through spatial transformer networks with a \gls*{tps} transformation, which models the distortion by finding and correcting fiducial points.
The second stage, \customEmph{Feature Extraction}, maps the input image to a representation that focuses on the attributes relevant to character recognition while suppressing irrelevant features such as font, color, size and background.
This task is usually performed by a module composed of \glspl*{cnn}, such as VGG, \acrshort*{resnet}, and \acrshort*{rcnn}.
The \customEmph{Sequence Modeling} stage converts visual features to contextual features that capture the context in the sequence of characters.
\gls*{bilstm} is generally employed for this task.
\ifextended
\else
Finally, the \customEmph{Prediction} stage produces the character sequence from the identified features.
This task is typically done by a \gls*{ctc} decoder or through an attention mechanism.
As can be seen in \cref{tab:related_work-miscellaneous:scene-text-recognition-methods-categorized}, while most methods can fit within this framework, they do not necessarily incorporate all four~modules.

\fi
\glsunset{rcnn}
\glsunset{resnet}

\begin{table}[!htb]
\vspace{-7mm}
\setlength{\tabcolsep}{5pt}
\centering
\caption{
Summary of seven well-known models for scene text recognition.
These models are listed chronologically and are further explored in other sections of this~work.%
}
\label{tab:related_work-miscellaneous:scene-text-recognition-methods-categorized}

\vspace{-2.75mm}

\resizebox{0.825\linewidth}{!}{%
\begin{tabular}{@{}lcccc@{}}
\toprule
Model    & Transformation & Feature Extraction & Sequence Modeling & Prediction \\ \midrule
\rtwoam~\cite{lee2016recursive}     & $-$              & \gls*{rcnn}               & $-$                 & Attention  \\
\rare~\cite{shi2016robust}     & \gls*{tps}            & VGG                & \gls*{bilstm}            & Attention  \\
\starnet~\cite{liu2016starnet} & \gls*{tps}            & \gls*{resnet}             & \gls*{bilstm}            & \gls*{ctc}        \\
\crnn~\cite{shi2017endtoend}     & $-$              & VGG                & \gls*{bilstm}            & \gls*{ctc}        \\
\grcnn~\cite{wang2017deep}    & $-$              & \gls*{rcnn}               & \gls*{bilstm}            & \gls*{ctc}        \\
\rosetta~\cite{borisyuk2018rosetta}  & $-$              & \gls*{resnet}             & $-$                 & \gls*{ctc}        \\
\trba~\cite{baek2019what}     & \gls*{tps}            & \gls*{resnet}             & \gls*{bilstm}            & Attention  \\ \bottomrule
\end{tabular}%
}

\end{table}

Atienza \cite{atienza2021vitstr} drew inspiration from the accomplishments of the \gls*{vit} and put forward a single-stage model named \vitstr for scene text recognition.
It operates by initially dividing the input image into non-overlapping patches.
These patches are then converted into $1$--D vector embeddings (i.e., flattened $2$--D patches).
To feed the encoder, each embedding is supplemented with a learnable patch embedding and a corresponding position~encoding.

Recent works on \gls*{lpr} have focused on developing multi-task \glspl*{cnn} that can process the entire \gls*{lp} image holistically, eliminating the need for character segmentation~\cite{spanhel2017holistic,goncalves2019multitask,fan2022improving}.
Two such models are \holistic~\cite{spanhel2017holistic} and \multitaskgabriel~\cite{goncalves2019multitask}.
In these models, the \gls*{lp} image undergoes initial processing via convolutional layers, followed by $N$ branches of fully connected layers.
Each branch is responsible for predicting a single character class (including a `blank' character) at a specific position on the \gls*{lp}, enabling the branches to collectively predict up to $N$ characters.
Both models are often used as baselines due to their remarkable balance between speed and accuracy~\cite{liu2021fast,gong2022unified,laroca2022cross,laroca2023do,nascimento2023super}.

The great speed/accuracy trade-off provided by YOLO networks~\cite{terven2023comprehensive} has inspired many authors to explore similar architectures targeting real-time performance for \gls*{lpr} and similar tasks.
Silva \& Jung~\cite{silva2020realtime} proposed \crnet, a YOLO-based model that effectively detects and recognizes all characters within a cropped \gls*{lp}~\cite{laroca2021efficient,laroca2022cross,silva2022flexible}.
Another noteworthy model is \fastocr~\cite{laroca2021towards}, which incorporates features from several object detectors that prioritize the trade-off between speed and accuracy.
In the domain of automatic meter reading~\cite{laroca2021towards}, \fastocr achieved considerably better results than multiple baselines that perform recognition holistically, including \crnn~\cite{shi2017endtoend}, \multitaskgabriel~\cite{goncalves2019multitask} and \trba~\cite{baek2019what}.

While we found a few works leveraging model fusion to improve \gls*{lpr} results, we observed that they explored a limited range of models and datasets in the experiments.
For example, Izidio et al.~\cite{izidio2020embedded} employed multiple instances of the same model (i.e., Tiny-YOLOv3) rather than different models with varying architectures.
Their experiments were conducted exclusively on a private dataset.
Another example is the very recent work by Schirrmacher et al.~\cite{schirrmacher2023benchmarking}, where they examined deep ensembles, BatchEnsemble, and Monte Carlo dropout using multiple instances of two backbone architectures.
The authors' primary focus was on recognizing severely degraded images, leading them to perform nearly all of their experiments on a synthetic dataset containing artificially degraded~images.

In summary, although the field of \gls*{lpr} has witnessed significant advancements through the development and application of deep learning-based models, there has been a noticeable lack of studies thoroughly examining the potential improvements in results by fusing the outputs from multiple recognition~models.
\section{Experimental Setup}
\label{sec:experimental-setup}

This section provides an overview of the setup adopted in our experiments.
We first enumerate the recognition models implemented for this study, providing specific information about the framework used for training and testing each of them, as well as the corresponding hyperparameters.
Subsequently, we compile a list of the datasets employed in our assessments, showcasing sample \gls*{lp} images from each dataset to highlight their diversity.
Afterward, we elaborate on the strategies examined for fusing the outputs of the different models.
Finally, we describe how the performance evaluation is carried~out.

The experiments were conducted on a PC with an AMD Ryzen Threadripper $1920$X $3.5$GHz CPU, $96$~GB of RAM operating at $2133$ MHz, an SSD~(read: $535$~MB/s; write: $445$~MB/s), and an NVIDIA Quadro RTX~$8000$ GPU ($48$~GB).

\subsection{Recognition Models}
\label{sec:experimental-setup:setup}

We explore $\numbaselines$ recognition models in our experiments:
\rare~\cite{shi2016robust}, \rtwoam~\cite{lee2016recursive}, \starnet~\cite{liu2016starnet}, \crnn~\cite{shi2017endtoend}, \grcnn~\cite{wang2017deep}, \holistic~\cite{spanhel2017holistic}, \multitaskgabriel~\cite{goncalves2019multitask}, \rosetta~\cite{borisyuk2018rosetta}, \trba~\cite{baek2019what}, \crnet~\cite{silva2020realtime}, \fastocr~\cite{laroca2021towards} and \vitstrbase~\cite{atienza2021vitstr}.
As discussed in \cref{sec:related_work}, these models were chosen because they rely on design patterns shared by many renowned models for scene text recognition, as well as for their frequent roles as baselines in recent \gls*{lpr} research~\cite{gong2022unified,laroca2022cross,kabiraj2023number,laroca2023do}.

We implemented each model using the original framework or well-known public repositories associated with it.
Specifically, we used Darknet\footnote{\mysize \url{https://github.com/AlexeyAB/darknet}} for the YOLO-based models (\crnet and \fastocr).
The multi-task models, \holistic and \multitaskgabriel, were trained and evaluated using Keras.
As for the remaining models, which were originally proposed for general scene text recognition, we used a fork\footnote{\mysize \url{https://github.com/roatienza/deep-text-recognition-benchmark/}} of the open source repository of Clova AI Research~(PyTorch).

Here we list the hyperparameters employed in each framework for training the recognition models.
These hyperparameters were determined based on existing research~\cite{baek2019what,atienza2021vitstr,laroca2022cross} and were further validated through experiments on the validation set.
In Darknet, the parameters include: \gls*{sgd} optimizer, $90$K iterations, a batch size of $64$, and a learning rate of [$10$\textsuperscript{-$3$},~$10$\textsuperscript{-$4$},~$10$\textsuperscript{-$5$}] with decay steps at $30$K and $60$K iterations.
In Keras, we employed the Adam optimizer with an initial learning rate of $10$\textsuperscript{-$3$} (ReduceLROnPlateau's patience of $5$ and factor of~$10$\textsuperscript{-$1$}), a batch size of $64$, and a patience value of $11$ (patience indicates the number of epochs without improvement before training is stopped).
In PyTorch, we used the following parameters: Adadelta optimizer with a decay rate of $\rho= 0.99$, $300$K iterations, and a batch size of~$128$.
The only modification we made to the models' architectures was adjusting the respective input layers to accommodate images with a width-to-height ratio of~$3$.

\subsection{Datasets}
\label{sec:experimental-setup:datasets}

Researchers have conducted experiments on various datasets to showcase the effectiveness of their systems in recognizing \glspl*{lp} from different regions~\cite{silva2018license,henry2020multinational,laroca2021efficient,lee2022license}.
As shown in \cref{tab:experiments:overview_datasets}, we perform experiments using images from \numdatasets public datasets commonly used to benchmark \gls*{alpr} systems~\cite{zeni2020weakly,silva2022flexible,laroca2022cross,laroca2022first,ke2023ultra}.
Each dataset was divided using standard splits, defined by the datasets' authors, or following previous works~\cite{laroca2021efficient,wang2022rethinking,ke2023ultra} (when there is no standard split)\footnote{\mysize Detailed information on which images were used to train, validate and test the models can be accessed at \urlSupplementary}.
Specifically, eight datasets were used both for training and evaluating the recognition models, while four were used exclusively for testing.
The selected datasets exhibit substantial diversity in terms of image quantity, acquisition settings, image resolution, and \gls*{lp} layouts.
As far as we know, no other work in \gls*{alpr} research has conducted experiments using images from such a wide range of public~datasets.

\begin{table}[!htb]
\centering
\setlength{\tabcolsep}{4pt}
\renewcommand{\arraystretch}{1.05}
\caption{
The datasets employed in our experimental analysis, with `$\ast$' indicating those used exclusively for testing (i.e., in cross-dataset experiments).
The ``Chinese'' layout denotes \glspl*{lp} assigned to vehicles registered in mainland China, while the ``Taiwanese'' layout corresponds to \glspl*{lp} issued for vehicles registered in the Taiwan~region.
}
\label{tab:experiments:overview_datasets}

\vspace{-2.5mm}

\resizebox{!}{8.65ex}{
\begin{tabular}{@{}lccc@{}}
\toprule
\textbf{Dataset} & \textbf{Year} & \phantom{-}\textbf{Images}\phantom{-} & \textbf{LP Layout} \\ \midrule
\caltech~\cite{caltech} & $1999$ & $126$ & American \\
\englishlp~\cite{englishlp} & $2003$ & $509$ & European \\
\stills~\cite{ucsd} & $2005$ & $291$ & American \\
\chineselp~\cite{zhou2012principal} & $2012$ & $411$ & Chinese \\
\aolp~\cite{hsu2013application} & $2013$ & $2{,}049$ & Taiwanese \\
OpenALPR-EU\normalsize{$^\ast$}\small~\cite{openalpreu} & $2016$ & $108$ & European \\ \bottomrule
\end{tabular}
} \vline \hspace{0.16mm}
\resizebox{!}{8.65ex}{
\begin{tabular}{@{}lccc@{}}
\toprule
\textbf{Dataset} & \textbf{Year} & \phantom{-}\textbf{Images}\phantom{-} & \textbf{LP Layout} \\ \midrule
\ssigsegplate~\cite{goncalves2016benchmark} & $2016$ & $2{,}000$ & Brazilian \\
PKU\normalsize{$^\ast$}\small~\cite{yuan2017robust} & $2017$ & $2{,}253$ & Chinese \\
\ufpralpr~\cite{laroca2018robust} & $2018$ & $4{,}500$ & Brazilian \\
CD-HARD\normalsize{$^\ast$}\small~\cite{silva2018license} & $2018$ & $102$ & Various \\
CLPD\normalsize{$^\ast$}\small~\cite{zhang2021robust_attentional} & $2021$ & $1{,}200$  & Chinese \\
\rodosolalpr~\cite{laroca2022cross} & $2022$ & $20{,}000$  & Brazilian \& Mercosur \\ \bottomrule
\end{tabular}
}
\end{table}

\begin{figure}[!htb]
    \centering
    \captionsetup[subfigure]{captionskip=2pt,justification=centering} 
    
    \resizebox{0.48\linewidth}{!}{
	\subfloat[][\caltech]{
    \includegraphics[height=6ex]{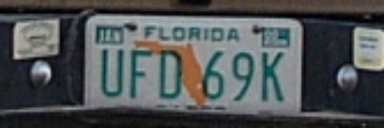}
    \includegraphics[height=6ex]{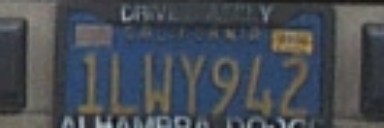}
    \includegraphics[height=6ex]{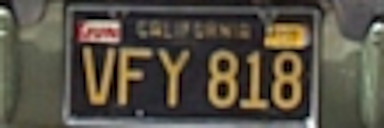}} 
    }
    \,
    \resizebox{0.48\linewidth}{!}{
	\subfloat[][\englishlp]{
    \includegraphics[height=6ex]{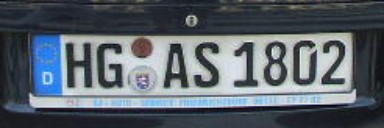}
    \includegraphics[height=6ex]{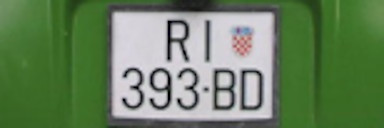}
    \includegraphics[height=6ex]{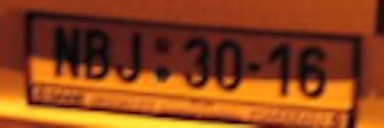}}  
    } \hspace{1mm}
    
    \vspace{1.25mm}
    
    \resizebox{0.48\linewidth}{!}{
	\subfloat[][\stills]{
    \includegraphics[height=6ex]{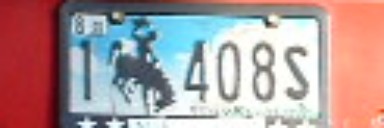}
    \includegraphics[height=6ex]{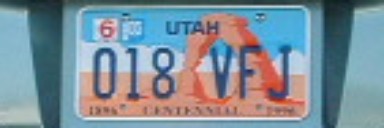}
    \includegraphics[height=6ex]{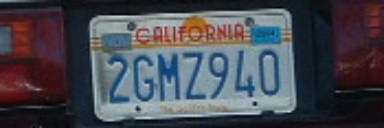}}
    }
    \,
    \resizebox{0.48\linewidth}{!}{
	\subfloat[][\chineselp\label{fig:samples-public-datasets-chineselp}]{
    \includegraphics[height=6ex]{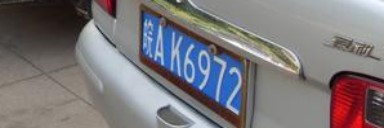}
    \includegraphics[height=6ex]{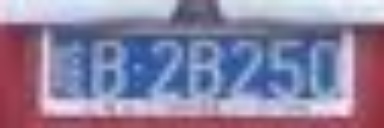}
    \includegraphics[height=6ex]{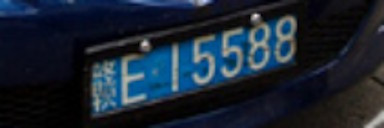}}  
    } \hspace{1mm}
    
    \vspace{1.25mm}
    
    \resizebox{0.48\linewidth}{!}{
	\subfloat[][\aolp]{
    \includegraphics[height=6ex]{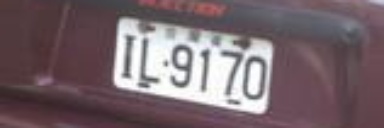}
    \includegraphics[height=6ex]{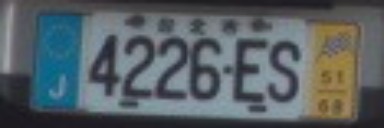}
    \includegraphics[height=6ex]{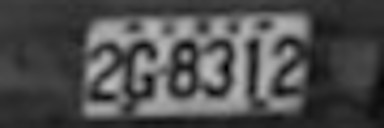}}
    }
    \,
    \resizebox{0.48\linewidth}{!}{
	\subfloat[][\openalpreu]{
    \includegraphics[height=6ex]{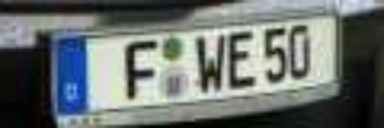}
    \includegraphics[height=6ex]{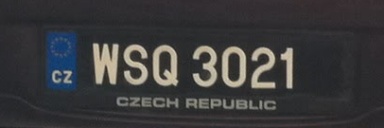}
    \includegraphics[height=6ex]{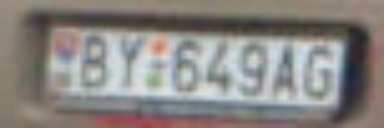}}  
    } \hspace{1mm}
    
    \vspace{1.25mm}
    
    \resizebox{0.48\linewidth}{!}{
	\subfloat[][\ssigsegplate]{
    \includegraphics[height=6ex]{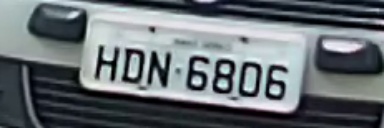}
    \includegraphics[height=6ex]{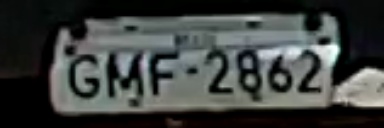}
    \includegraphics[height=6ex]{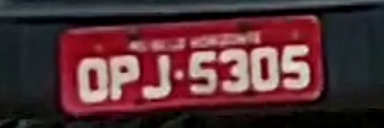}}
    }
    \,
    \resizebox{0.48\linewidth}{!}{
	\subfloat[][\pku]{
    \includegraphics[height=6ex]{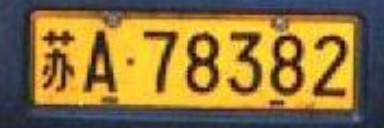}
    \includegraphics[height=6ex]{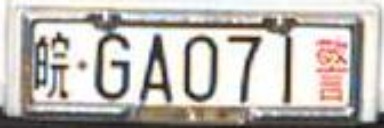}
    \includegraphics[height=6ex]{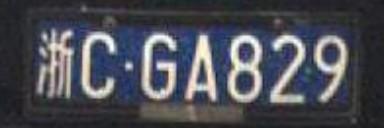}} 
    } \hspace{1mm} 
    
    \vspace{1.25mm}
    
    \resizebox{0.48\linewidth}{!}{
	\subfloat[][\ufpralpr]{
    \includegraphics[height=6ex]{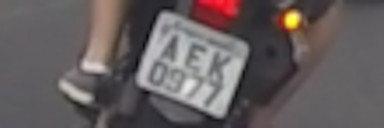}
    \includegraphics[height=6ex]{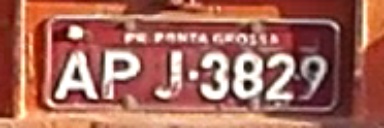}
    \includegraphics[height=6ex]{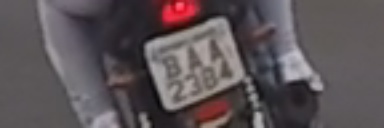}} 
    }
    \,
    \resizebox{0.48\linewidth}{!}{
	\subfloat[][\cdhard]{
    \includegraphics[height=6ex]{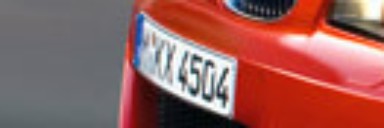}
    \includegraphics[height=6ex]{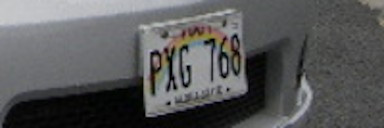}
    \includegraphics[height=6ex]{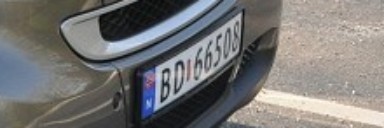}} 
    } \hspace{1mm} 

    \vspace{1.25mm}
    
    \resizebox{0.48\linewidth}{!}{
	\subfloat[][\clpd]{
    \includegraphics[height=6ex]{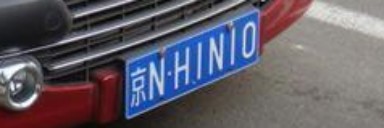}
    \includegraphics[height=6ex]{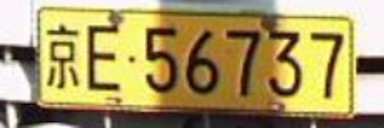}
    \includegraphics[height=6ex]{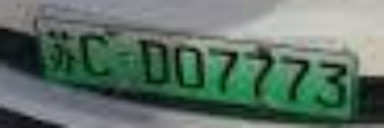}} 
    }
    \,
    \resizebox{0.48\linewidth}{!}{
	\subfloat[][\rodosolalpr]{
    \includegraphics[height=5.9ex]{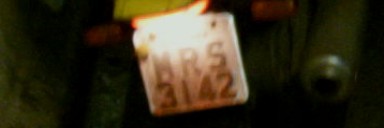}
    \includegraphics[height=5.9ex]{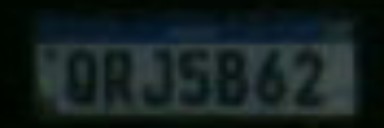}
    \includegraphics[height=5.9ex]{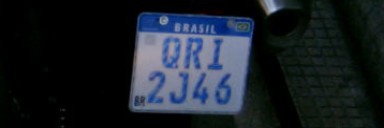}} 
    } \hspace{1mm}

    \vspace{-2mm}
    
    \caption{Some \gls*{lp} images from the public datasets used in our experimental evaluation.
    }
    \label{fig:samples-public-datasets}
\end{figure}

The diversity of \gls*{lp} layouts across the selected datasets is depicted in \cref{fig:samples-public-datasets}, revealing considerable variations even among \glspl*{lp} from the same region.
For instance, the \englishlp and \openalpreu datasets, both collected in Europe, include images of \glspl*{lp} with notable distinctions in colors, aspect ratios, symbols (e.g., coats of arms), and the number of characters.
Furthermore, certain datasets encompass \glspl*{lp} with two rows of characters, shadows, tilted orientations, and at relatively low spatial~resolutions.

\begin{figure}[!htb]
    \vspace{1.75mm}
    \centering
    
    \resizebox{0.995\linewidth}{!}{
    \resizebox{0.48\linewidth}{!}{
    \includegraphics[width=0.19\linewidth]{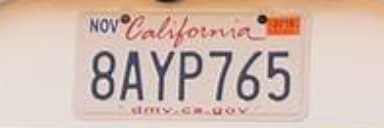} \hspace{-1.4mm}
    \includegraphics[width=0.19\linewidth]{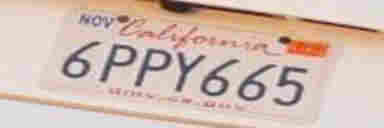} \hspace{-1.4mm}
    \includegraphics[width=0.19\linewidth]{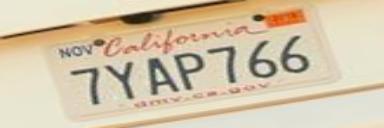} \hspace{-1.4mm}
    \includegraphics[width=0.19\linewidth]{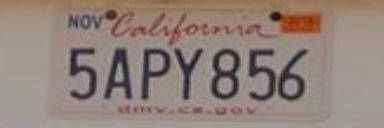}
    } 
    \resizebox{0.48\linewidth}{!}{    
    \includegraphics[width=0.19\linewidth]{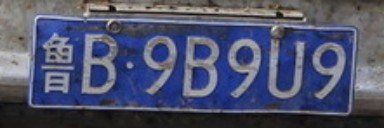} \hspace{-1.4mm}
    \includegraphics[width=0.19\linewidth]{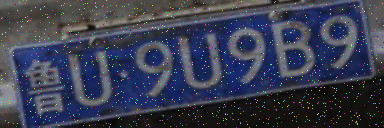} \hspace{-1.4mm}
    \includegraphics[width=0.19\linewidth]{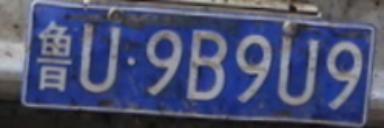} \hspace{-1.4mm}
    \includegraphics[width=0.19\linewidth]{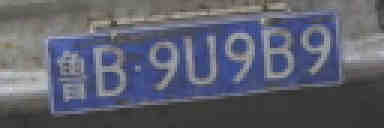}
    } 
    }

    \vspace{0.3mm}
    
    \resizebox{0.995\linewidth}{!}{
    \resizebox{0.48\linewidth}{!}{
    \includegraphics[width=0.19\linewidth]{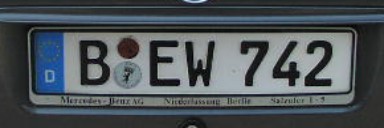} \hspace{-1.4mm}
    \includegraphics[width=0.19\linewidth]{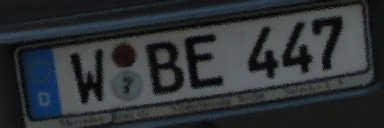} \hspace{-1.4mm}
    \includegraphics[width=0.19\linewidth]{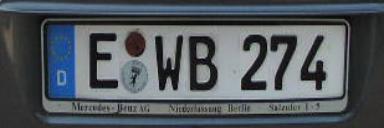} \hspace{-1.4mm}
    \includegraphics[width=0.19\linewidth]{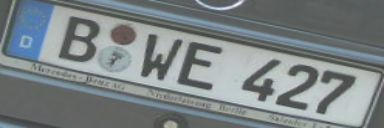}
    }
    \resizebox{0.48\linewidth}{!}{
    \includegraphics[width=0.19\linewidth]{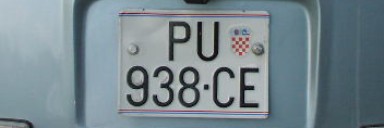} \hspace{-1.4mm}
    \includegraphics[width=0.19\linewidth]{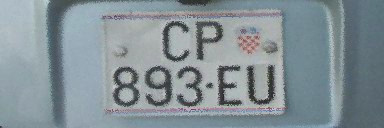} \hspace{-1.4mm}
    \includegraphics[width=0.19\linewidth]{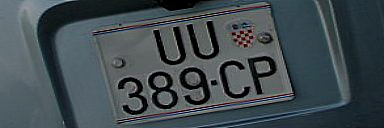} \hspace{-1.4mm}
    \includegraphics[width=0.19\linewidth]{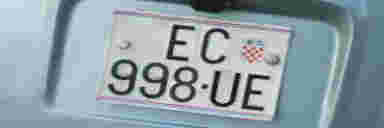}
    }
    }

    \vspace{-2.75mm}
    
    \caption{
    Examples of \gls*{lp} images we created to mitigate overfitting.
    Within each group, the image on the left is the original, while the remaining ones are artificially generated counterparts.
    Various transformations were applied to enhance image~variability.
    }
    \label{fig:data-augmentation-samples-permutation}

    \vspace{-1.75mm}
\end{figure}

We explored various data augmentation techniques to ensure a balanced distribution of training images across different datasets. 
These techniques include random cropping, the introduction of random shadows, grayscale conversion, and random perturbations of hue, saturation, and brightness.
Additionally, to counteract the propensity of recognition models to memorize sequence patterns encountered during training~\cite{goncalves2018realtime,zeni2020weakly,garcia2022out}, we generated many synthetic \gls*{lp} images by shuffling the character positions on each \gls*{lp} (using the labels provided in~\cite{laroca2021efficient}).
Examples of these generated images are shown in~\cref{fig:data-augmentation-samples-permutation}.

To further mitigate the inherent biases present in public datasets~\cite{laroca2022first}, we expanded our training set by including $772$ images from the internet.
These images were annotated and made available by Ref.~\cite{laroca2021efficient}. 
This supplementary dataset comprises $257$ American \glspl*{lp}, $347$ Chinese \glspl*{lp}, and $178$ European~\glspl*{lp}.

\subsection{Fusion Approaches}
\label{sec:experimental-setup:fusion}

This study examines three primary approaches to combine the outputs of multiple recognition models.
The first approach involves selecting the sequence predicted with the \customEmph{\gls*{hc}} value as the final prediction, even if only one model predicts it.
The second approach employs the \customEmph{\gls*{mv}} rule to aggregate the sequences predicted by the different models.
In other words, the final prediction is the sequence predicted by the largest number of models, disregarding the confidence values associated with each prediction.
Lastly, the third approach follows a similar \customEmph{Majority Vote} rule but performs individual aggregation for each \customEmph{Character Position}~(\acrshort*{mvcp}).
To illustrate, the characters predicted in the first position are analyzed separately, and the character predicted the most times is selected.
The same process is then applied to each subsequent character position until the last one.
Ultimately, the selected characters are concatenated to form the final~string.

\glsunset{mvcp}

One concern that arises when employing majority vote-based strategies is the potential occurrence of a tie.
Let's consider a scenario where an \gls*{lp} image is processed by five recognition models.
Two models predict ``ABC-123,'' two models predict ``ABC-124,'' and the remaining model predicts ``ABC-125.''
In this case, a tie occurs between ``ABC-123'' and ``ABC-124.''
To address this, we assess two tie-breaking approaches for each majority vote strategy: 
(i)~selecting the prediction made with the highest confidence among the tied predictions as the final one, and (ii)~selecting the prediction made by the ``best model'' as the final prediction.
In this study, for simplicity, we consider the best model the one that performs best individually across all datasets.
However, in a more practical scenario, the chosen model could be the one known to perform best in the specific implementation scenario (e.g., one model may be the most robust for recognizing tilted \glspl*{lp} while another model may excel at handling low-resolution or noisy~images).
We use the acronym \mvhc to refer to the majority vote approach in which ties are broken by selecting the prediction made with the highest confidence value.
Similarly, \mvbm refers to the majority vote approach in which ties are resolved by choosing the prediction made by the best model.
The \mvcp approaches follow a similar naming convention (\mvcp--HC and~\mvcp--BM).

It is important to mention that when conducting fusion based on the highest confidence, we consider the confidence values derived directly from the models' outputs, even though some of them tend to make overconfident predictions.
We carried out several experiments in which we normalized the confidence values of different models before fusing them, using various strategies such as weighted normalization based on the average confidence of each classifier's predictions.
Somewhat surprisingly, these attempts did not yield improved~results.

\subsection{Performance Evaluation}

In line with the standard practice in the literature, we report the performance of each experiment by calculating the ratio of correctly recognized \glspl*{lp} to the total number of \glspl*{lp} in the test set.
An \gls*{lp} is considered correctly recognized only if all the characters on the \gls*{lp} are accurately identified, as even a single incorrectly recognized character can lead to the misidentification of the vehicle.

It is important to note that, although this work focuses on the \gls*{lpr} stage, the \gls*{lp} images used as input for the recognition models were not directly cropped from the ground truth.
Instead, the YOLOv4 model~\cite{bochkovskiy2020yolov4} was employed to detect the \glspl*{lp}.
This approach allows for a more accurate simulation of real-world scenarios, considering the imperfect nature of \gls*{lp} detection and the reduced robustness of certain recognition models when faced with imprecisely detected \gls*{lp} regions~\cite{goncalves2018realtime,lee2022license}.
As in~\cite{laroca2022cross}, the results obtained using YOLOv4 were highly satisfactory.
Considering detections with an \gls*{iou} $\ge0.7$ as correct, YOLOv4 achieved an average recall rate exceeding $99.7$\% in the test sets of the datasets used for training and validation, and $97.8$\% in the cross-dataset experiments.
In both cases, the precision rates obtained were greater than~$97$\%.
\section{Results}
\label{sec:results}

\cref{tab:results-trad} shows the recognition rates obtained on the disjoint test sets of the \numdatasetsTradWords datasets used for training and validating the models.
It presents the results reached by each model individually, as well as the outcomes achieved through the fusion strategies outlined in \cref{sec:experimental-setup:fusion}.
To improve clarity, \cref{tab:results-trad} only includes the best results attained through model fusion.
For a detailed breakdown of the results achieved by combining the outputs from the top $2$ to the top $12$ recognition models, refer to \cref{tab:ranking-models}.
The ranking of the models was determined based on their mean performance across the~datasets (the ranking on the validation set was essentially the same, with only two models swapping~positions).

\begin{table*}[!htb]
\centering
\renewcommand{\arraystretch}{1.05}
\setlength{\tabcolsep}{4pt}
\caption{
Comparison of the recognition rates achieved across \numdatasetsTradWords popular datasets by \nummodels models individually and through five different fusion strategies.
Each model (rows) was trained once on the combined set of training images from all datasets and evaluated on the respective test sets (columns).
The models are listed alphabetically, and the best recognition rates achieved in each dataset are shown in~bold.
}
\label{tab:results-trad}

\vspace{-2mm}

\newcommand{\as}[1] {\fontsize{10.33}{12.77} #1} %

\resizebox{0.99\textwidth}{!}{%
\begin{tabular}{@{}lccccccccc@{}}
\toprule
\diagbox[trim=l,innerrightsep=14pt]
{Approach}{Test set}    & \multicolumn{1}{c}{\begin{tabular}[c]{@{}c@{}}\caltech\\\# $46$\end{tabular}} & \multicolumn{1}{c}{\begin{tabular}[c]{@{}c@{}}\englishlp\\\# $102$\end{tabular}} & \multicolumn{1}{c}{\begin{tabular}[c]{@{}c@{}}\stills\\\# $60$\end{tabular}} & \multicolumn{1}{c}{\begin{tabular}[c]{@{}c@{}}\chineselp\\\# $161$\end{tabular}} & \multicolumn{1}{c}{\begin{tabular}[c]{@{}c@{}}\aolp\\\# $687$\end{tabular}}  & \multicolumn{1}{c}{\begin{tabular}[c]{@{}c@{}}\ssigsegplate\\\# $804$\end{tabular}} & \multicolumn{1}{c}{\begin{tabular}[c]{@{}c@{}}\ufpralpr\\\# $1{,}800$\end{tabular}} & \multicolumn{1}{c}{\begin{tabular}[c]{@{}c@{}}\rodosolalpr\\\# $8{,}000$\end{tabular}} & Average \\ \midrule
\crnet~\cite{silva2020realtime}    & \as{$\textbf{97.8}$\textbf{\%}}      & \phantom{---}\as{$94.1$\%}\phantom{---}   & \phantom{--}\as{$\textbf{100.0}$\textbf{\%}}\phantom{--}     & \phantom{---}\as{$\textbf{97.5}$\textbf{\%}}\phantom{---}   & \phantom{---}\as{$98.1$\%}\phantom{---}     & \as{$\textbf{97.5}$\textbf{\%}}       & \as{$82.6$\%}     & \phantom{$^\dagger$}\as{$59.0$\%}$^\dagger$      & \as{$90.8$\%} \\
\crnn~\cite{shi2017endtoend}        & \as{$93.5$\%}      & \as{$88.2$\%}   & \as{$91.7$\%}     & \as{$90.7$\%}   & \as{$97.1$\%}      & \as{$92.9$\%}       & \as{$68.9$\%}       & \as{$73.6$\%}      & \as{$87.1$\%} \\
\fastocr~\cite{laroca2021towards}    & \as{$93.5$\%}      & \as{$\textbf{97.1}$\textbf{\%}}   & \as{$\textbf{100.0}$\textbf{\%}}     & \as{$\textbf{97.5}$\textbf{\%}}   & \as{$98.1$\%}     & \as{$97.1$\%}       & \as{$81.6$\%}       & \phantom{$^\dagger$}\as{$56.7$\%}$^\dagger$      & \as{$90.2$\%} \\
\grcnn~\cite{wang2017deep}       & \as{$93.5$\%}      & \as{$92.2$\%}   & \as{$93.3$\%}     & \as{$91.9$\%}   & \as{$97.1$\%}     & \as{$93.4$\%}       & \as{$66.6$\%}      & \as{$77.6$\%}      & \as{$88.2$\%} \\
\holistic~\cite{spanhel2017holistic}     & \as{$87.0$\%}      & \as{$75.5$\%}   & \as{$88.3$\%}     & \as{$95.0$\%}   & \as{$97.7$\%}     & \as{$95.6$\%}       & \as{$81.2$\%}       & \as{$94.7$\%}      & \as{$89.4$\%} \\
\multitaskgabriel~\cite{goncalves2019multitask}        & \as{$89.1$\%}      & \as{$73.5$\%}   & \as{$85.0$\%}     & \as{$92.5$\%}   & \as{$94.9$\%}      & \as{$93.3$\%}       & \as{$72.3$\%}       & \as{$86.6$\%}      & \as{$85.9$\%} \\
\rtwoam~\cite{lee2016recursive}        & \as{$89.1$\%}      & \as{$83.3$\%}   & \as{$86.7$\%}     & \as{$91.9$\%}   & \as{$96.5$\%}     & \as{$92.0$\%}       & \as{$75.9$\%}       & \as{$83.4$\%}      & \as{$87.4$\%} \\
\rare~\cite{shi2016robust}        & \as{$95.7$\%}      & \as{$94.1$\%}   & \as{$95.0$\%}     & \as{$94.4$\%}   & \as{$97.7$\%}     & \as{$94.0$\%}       & \as{$75.7$\%}       & \as{$78.7$\%}      & \as{$90.7$\%} \\
\rosetta~\cite{borisyuk2018rosetta}     & \as{$89.1$\%}      & \as{$82.4$\%}   & \as{$93.3$\%}     & \as{$93.8$\%}   & \as{$97.5$\%}     & \as{$94.4$\%}       & \as{$75.5$\%}       & \as{$89.0$\%}      & \as{$89.4$\%} \\
\starnet~\cite{liu2016starnet}    & \as{$95.7$\%}      & \as{$96.1$\%}   & \as{$95.0$\%}     & \as{$95.7$\%}   & \as{$97.8$\%}      & \as{$96.1$\%}       & \as{$78.8$\%}       & \as{$82.3$\%}      & \as{$92.2$\%} \\
\trba~\cite{baek2019what}        & \as{$93.5$\%}      & \as{$91.2$\%}   & \as{$91.7$\%}     & \as{$93.8$\%}   & \as{$97.2$\%}     & \as{$97.3$\%}       & \as{$83.4$\%}       & \as{$80.6$\%}     & \as{$91.1$\%} \\
\vitstrbase~\cite{atienza2021vitstr} & \as{$87.0$\%}      & \as{$88.2$\%}   & \as{$86.7$\%}     & \as{$96.9$\%}   & \as{$\textbf{99.4}$\textbf{\%}}      & \as{$95.8$\%}       & \as{$\textbf{89.7}$\textbf{\%}}       & \as{$\textbf{95.6}$\textbf{\%}}      & \as{\phantom{a}$\textbf{92.4}$\textbf{\%}} \\[3pt] \cdashline{1-10} \\[-9pt] \cdashline{1-10} \\[-6pt]
Fusion HC (\textit{top 6})    & \as{$\textbf{\accAvgHcbmCaltech}$\textbf{\%}}      & \as{$\accAvgHcbmEnglishlp$\%}   & \as{$\accAvgHcbmStills$\%}     & \as{$\textbf{\accAvgHcbmChineselp}$\textbf{\%}}   & \as{$\accAvgHcbmAolp$\%}     & \as{$\accAvgHcbmSsig$\%}       & \as{$\accAvgHcbmUfpr$\%}       & \as{$\accAvgHcbmRodosol$\%}      & \as{$\accAvgHcbmTrad$\%} \\
Fusion MV--BM (\textit{top 8})    & \as{$\textbf{\accAvgMvbmCaltech}$\textbf{\%}}      & \as{$\textbf{\accAvgMvbmEnglishlp}$\textbf{\%}}   & \as{$\textbf{\accAvgMvbmStills}$\textbf{\%}}     & \as{$\textbf{\accAvgMvbmChineselp}$\textbf{\%}}   & \as{$\textbf{\accAvgMvbmAolp}$\textbf{\%}}     & \as{$\accAvgMvbmSsig$\%}       & \as{$\accAvgMvbmUfpr$\%}       & \as{$\accAvgMvbmRodosol$\%}      & \as{$\accAvgMvbmTrad$\%} \\
Fusion MV--HC (\textit{top 8})    & \as{$\textbf{\accAvgMvhcCaltech}$\textbf{\%}}      & \as{$\textbf{\accAvgMvhcEnglishlp}$\textbf{\%}}   & \as{$\textbf{\accAvgMvhcStills}$\textbf{\%}}     & \as{$\textbf{\accAvgMvhcChineselp}$\textbf{\%}}   & \as{$\textbf{\accAvgMvhcAolp}$\textbf{\%}}     & \as{$\accAvgMvhcSsig$\%}       & \as{$\accAvgMvhcUfpr$\%}       & \as{$\textbf{\accAvgMvhcRodosol}$\textbf{\%}}      & \as{$\textbf{\accAvgMvhcTrad}$\textbf{\%}} \\
Fusion MVCP--BM (\textit{top 9})    & \as{$\accAvgMvcpBmCaltech$\%}      & \as{$\accAvgMvcpBmEnglishlp$\%}   & \as{$\textbf{\accAvgMvcpBmStills}$\textbf{\%}}     & \as{$\textbf{\accAvgMvcpBmChineselp}$\textbf{\%}}   & \as{$\accAvgMvcpBmAolp$\%}     & \as{$\accAvgMvcpBmSsig$\%}       & \as{$\textbf{\accAvgMvcpBmUfpr}$\textbf{\%}}       & \as{$\accAvgMvcpBmRodosol$\%}      & \as{$\accAvgMvcpBmTrad$\%} \\
Fusion MVCP--HC (\textit{top 9})    & \as{$\textbf{\accAvgMvcpHcCaltech}$\textbf{\%}}      & \as{$\accAvgMvcpHcEnglishlp$\%}   & \as{$\textbf{\accAvgMvcpHcStills}$\textbf{\%}}     & \as{$\textbf{\accAvgMvcpHcChineselp}$\textbf{\%}}   & \as{$\accAvgMvcpHcAolp$\%}     & \as{$\textbf{\accAvgMvcpHcSsig}$\textbf{\%}}       & \as{$\accAvgMvcpHcUfpr$\textbf{\%}}       & \as{$\accAvgMvcpHcRodosol$\%}      & \as{$\accAvgMvcpHcTrad$\%} \\
\bottomrule \\[-2.3ex]
\multicolumn{10}{l}{\resizebox{1.825\linewidth}{!}{ \hspace{-2mm}\as{$^{\dagger}$}\hspace{0.3mm}Images from the \rodosolalpr dataset were not used for training the \crnet and \fastocr models, as each character’s bounding box needs to be labeled for training~them.}}
\end{tabular}%
}
\end{table*}

\begin{table}[!htb]
    \centering
    \setlength{\tabcolsep}{5pt}
    \caption{
    Average results obtained across the datasets by combining the output of the top $N$ recognition models, ranked by accuracy, using five distinct~strategies.%
    }
    \label{tab:ranking-models}
    
    \vspace{-2mm}

    \resizebox{0.85\linewidth}{!}{
    \begin{tabular}{@{}lccccc@{}}
    \toprule
     Models & HC & \phantom{ii}MV--BM\phantom{ii} & \phantom{i}MV--HC\phantom{i} & MVCP--BM & MVCP--HC \\ \midrule
    Top $1$ (\vitstrbase) & \phantom{ii}92.4\%\phantom{ii} & $92.4$\% & $92.4$\% & $92.4$\% & $92.4$\% \\
    Top $2$ (+ \starnet) & $94.1$\% & $92.4$\% & $94.1$\% & $92.4$\% & $94.1$\% \\
    Top $3$ (+ \trba) & $94.2$\% & $94.6$\% & $94.9$\% & $94.2$\% & $94.2$\% \\
    Top $4$ (+ \crnet) & $95.2$\% & $95.9$\% & $96.3$\% & $94.8$\% & $95.9$\% \\
    Top $5$ (+ \rare) & $95.5$\% & $96.1$\% & $96.6$\% & $96.1$\% & $96.2$\% \\
    Top $6$ (+ \fastocr) & $\textbf{96.0}$\textbf{\%} & $97.1$\% & $97.0$\% & $96.7$\% & $96.9$\% \\
    Top $7$ (+ \rosetta) & $95.4$\% & $97.3$\% & $97.2$\% & $97.1$\% & $97.0$\% \\
    Top $8$ (+ \holistic) & $95.7$\% & $\textbf{97.5}$\textbf{\%} & $\textbf{97.6}$\textbf{\%} & $96.1$\% & $97.2$\% \\
    Top $9$ (+ \grcnn) & $95.7$\% & $97.5$\% & $97.5$\% & $\textbf{97.2}$\textbf{\%} & $\textbf{97.5}$\textbf{\%} \\
    Top $10$ (+ \rtwoam) & $95.5$\% & $97.4$\% & $97.2$\% & $96.1$\% & $96.6$\% \\
    Top $11$ (+ \crnn) & $95.2$\% & $97.1$\% & $97.0$\% & $96.5$\% & $96.5$\% \\
    Top $12$ (+ \multitaskgabriel) & $95.0$\% & $97.0$\% & $97.0$\% & $95.5$\% & $96.5$\% \\ \bottomrule
    \end{tabular}%
    }
\end{table}

Upon analyzing the results presented in \cref{tab:results-trad}, it becomes evident that model fusion has yielded substantial improvements.
Specifically, the highest average recognition rate increased from \bestIndividual (\vitstrbase) to \bestFusion by combining the outputs of multiple recognition models (\mvhc).
While each model individually obtained recognition rates below $90$\% for at least two datasets (three on average), all fusion strategies surpassed the $90$\% threshold across all datasets.
Remarkably, in most cases, fusion led to recognition rates exceeding $95$\%.

The significance of conducting experiments on multiple datasets becomes apparent as we observe that the best overall model (\vitstrbase) did not achieve the top result in five of the eight datasets.
Notably, it exhibited relatively poor performance on the \caltech, \englishlp, and \stills datasets.
We attribute this to two primary reasons: (i)~these datasets are older, containing fewer training images, which seems to impact certain models more than others (as explained in \cref{sec:experimental-setup:datasets}, we exploited data augmentation techniques to mitigate this issue); and (ii)~these datasets were collected in the United States and Europe, regions known for having a higher degree of variability in \gls*{lp} layouts compared to the regions where the other datasets were collected (specifically, Brazil, mainland China, and Taiwan).
It is worth noting that we included these datasets in our experimental setup, despite their limited number of images, precisely because they provide an opportunity to uncover such valuable~insights.

\begin{figure}[!htb]
	\centering
	\captionsetup[subfigure]{labelformat=empty,font={footnotesize}}

	\resizebox{0.99\linewidth}{!}{
		\subfloat[][\centering \textbf{\phantom{i}\phantom{1}\texttt{\vitstrbase}:} \texttt{AIQ1\red{Q}56 (0.93)}\hspace{\textwidth}\textbf{\phantom{i}\texttt{\phantom{1111}\starnet}:} \texttt{A\red{T}Q1056 (0.59)}\hspace{\textwidth}\textbf{\phantom{i}\texttt{\phantom{11111111}\trba}:} \texttt{AIQ1056 (0.98)}\hspace{\textwidth}\textbf{\phantom{i}\texttt{\phantom{111111}\crnet}:} \texttt{AIQ1056 (0.82)}\hspace{\textwidth}\textbf{\phantom{i}\texttt{\phantom{11111111}\rare}:} \texttt{AIQ1\red{Q}56 (0.92)}\hspace{\textwidth}\textbf{\phantom{i}\texttt{Fusion MV--HC}:} \texttt{AIQ1056\phantom{ (0.00)}}]{
			\includegraphics[width=0.39\linewidth]{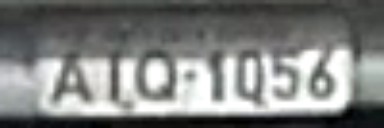}} \,
		\subfloat[][\centering \textbf{\phantom{i}\phantom{1}\texttt{\vitstrbase}:} \texttt{AS5I8D (0.53)}\hspace{\textwidth}\textbf{\phantom{i}\texttt{\phantom{1111}\starnet}:} \texttt{AS518\red{0} (0.82)}\hspace{\textwidth}\textbf{\phantom{i}\texttt{\phantom{11111111}\trba}:} \texttt{AS518\red{0} (0.60)}\hspace{\textwidth}\textbf{\phantom{i}\texttt{\phantom{111111}\crnet}:} \texttt{AS518D (0.83)}\hspace{\textwidth}\textbf{\phantom{i}\texttt{\phantom{11111111}\rare}:} \texttt{AS5I8D (0.79)}\hspace{\textwidth}\textbf{\phantom{i}\texttt{Fusion MV--HC}:} \texttt{AS5I8D\phantom{ (0.00)}}]{
			\includegraphics[width=0.39\linewidth]{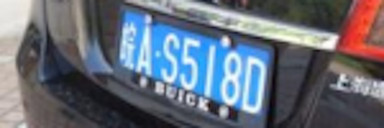}} \,
		\subfloat[][\centering \textbf{\phantom{i}\phantom{1}\texttt{\vitstrbase}:} \texttt{4NIU770 (0.45)}\hspace{\textwidth}\textbf{\phantom{i}\texttt{\phantom{1111}\starnet}:} \texttt{4NIU770 (0.94)}\hspace{\textwidth}\textbf{\phantom{i}\texttt{\phantom{11111111}\trba}:} \texttt{4N\red{T}U770 (0.99)}\hspace{\textwidth}\textbf{\phantom{i}\texttt{\phantom{111111}\crnet}:} \texttt{4N\red{T}U770 (0.91)}\hspace{\textwidth}\textbf{\phantom{i}\texttt{\phantom{11111111}\rare}:} \texttt{4NIU770 (0.99)}\hspace{\textwidth}\textbf{\phantom{i}\texttt{Fusion MV--HC}:} \texttt{4NIU770\phantom{ (0.00)}}]{
			\includegraphics[width=0.39\linewidth]{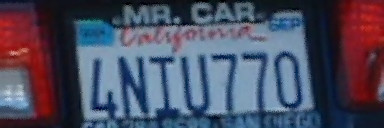}} \,
		\subfloat[][\centering \textbf{\phantom{i}\phantom{1}\texttt{\vitstrbase}:} \texttt{\red{5}EZ\red{Z}29 (0.51)}\hspace{\textwidth}\textbf{\phantom{i}\texttt{\phantom{1111}\starnet}:} \texttt{SEZ229 (0.74)}\hspace{\textwidth}\textbf{\phantom{i}\texttt{\phantom{11111111}\trba}:} \texttt{\red{5}EZ229 (0.99)}\hspace{\textwidth}\textbf{\phantom{i}\texttt{\phantom{111111}\crnet}:} \texttt{\red{5}EZ229 (0.88)}\hspace{\textwidth}\textbf{\phantom{i}\texttt{\phantom{11111111}\rare}:} \texttt{\red{5}EZ229 (0.88)}\hspace{\textwidth}\textbf{\phantom{i}\texttt{Fusion MV--HC}:} \texttt{\red{5}EZ229\phantom{ (0.00)}}]{
			\includegraphics[width=0.39\linewidth]{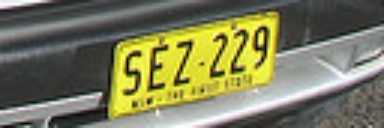}} \,
	} 
	
	\vspace{3mm}
 
	\resizebox{0.99\linewidth}{!}{
		\subfloat[][\centering \textbf{\phantom{i}\phantom{1}\texttt{\vitstrbase}:} \texttt{KRM7E95 (0.99)}\hspace{\textwidth}\textbf{\phantom{i}\texttt{\phantom{1111}\starnet}:} \texttt{KR\red{H}7E95 (0.59)}\hspace{\textwidth}\textbf{\phantom{i}\texttt{\phantom{11111111}\trba}:} \texttt{KRM7E95 (0.51)}\hspace{\textwidth}\textbf{\phantom{i}\texttt{\phantom{111111}\crnet}:} \texttt{KR\red{H}7E95 (0.73)}\hspace{\textwidth}\textbf{\phantom{i}\texttt{\phantom{11111111}\rare}:} \texttt{KRM7E95 (0.60)}\hspace{\textwidth}\textbf{\phantom{i}\texttt{Fusion MV--HC}:} \texttt{KRM7E95\phantom{ (0.00)}}]{
			\includegraphics[width=0.39\linewidth]{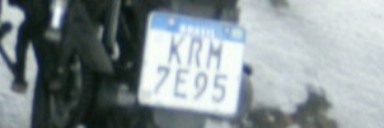}} \,
		\subfloat[][\centering \textbf{\phantom{i}\phantom{1}\texttt{\vitstrbase}:} \texttt{Y88096 (0.94)}\hspace{\textwidth}\textbf{\phantom{i}\texttt{\phantom{1111}\starnet}:} \texttt{Y\red{6}8096 (0.93)}\hspace{\textwidth}\textbf{\phantom{i}\texttt{\phantom{11111111}\trba}:} \texttt{Y88096 (0.97)}\hspace{\textwidth}\textbf{\phantom{i}\texttt{\phantom{111111}\crnet}:} \texttt{Y\red{96}096 (0.75)}\hspace{\textwidth}\textbf{\phantom{i}\texttt{\phantom{11111111}\rare}:} \texttt{Y\red{S}8096 (0.67)}\hspace{\textwidth}\textbf{\phantom{i}\texttt{Fusion MV--HC}:} \texttt{Y88096\phantom{ (0.00)}}]{
			\includegraphics[width=0.39\linewidth]{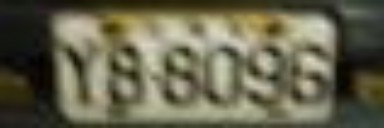}} \,
		\subfloat[][\centering \textbf{\phantom{i}\phantom{1}\texttt{\vitstrbase}:} \texttt{HLP459\red{A} (0.98)}\hspace{\textwidth}\textbf{\phantom{i}\texttt{\phantom{1111}\starnet}:} \texttt{HLP4594 (0.97)}\hspace{\textwidth}\textbf{\phantom{i}\texttt{\phantom{11111111}\trba}:} \texttt{HLP\red{A}594 (0.99)}\hspace{\textwidth}\textbf{\phantom{i}\texttt{\phantom{111111}\crnet}:} \texttt{HLP4594 (0.85)}\hspace{\textwidth}\textbf{\phantom{i}\texttt{\phantom{11111111}\rare}:} \texttt{HLP\red{A}59\red{A} (0.93)}\hspace{\textwidth}\textbf{\phantom{i}\texttt{Fusion MV--HC}:} \texttt{HLP4594\phantom{ (0.00)}}]{
			\includegraphics[width=0.39\linewidth]{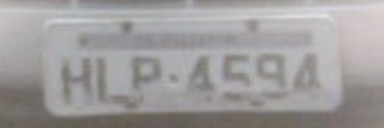}} \,
		\subfloat[][\centering \textbf{\phantom{i}\phantom{1}\texttt{\vitstrbase}:} \texttt{MRU3095 (0.97)}\hspace{\textwidth}\textbf{\phantom{i}\texttt{\phantom{1111}\starnet}:} \texttt{MR\red{0}3095 (0.98)}\hspace{\textwidth}\textbf{\phantom{i}\texttt{\phantom{11111111}\trba}:} \texttt{MR\red{D}3095 (0.72)}\hspace{\textwidth}\textbf{\phantom{i}\texttt{\phantom{111111}\crnet}:} \texttt{MR\red{D}3095 (0.94)}\hspace{\textwidth}\textbf{\phantom{i}\texttt{\phantom{11111111}\rare}:} \texttt{MR\red{D}3095 (0.87)}\hspace{\textwidth}\textbf{\phantom{i}\texttt{Fusion MV--HC}:} \texttt{MR\red{D}3095\phantom{ (0.00)}}]{
			\includegraphics[width=0.39\linewidth]{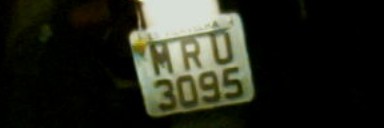}} \,
	}
	
	  \vspace{-0.75mm}
	    
	\caption{
        Predictions obtained in eight \gls*{lp} images by multiple models individually and through the best fusion approach.
        Although we only show the predictions from the top~$5$ models for better viewing, it is noteworthy that in these particular cases, fusing the top~$8$ models (the optimal configuration) yielded identical predictions.
        The confidence for each prediction is indicated in parentheses, and any errors are highlighted in~red.
        }
	\label{fig:qualitative-model-fusion}
\end{figure}

Basically, by analyzing the results reported for each dataset individually, we observe that combining the outputs of multiple models does not necessarily lead to significantly improved performance compared to the best model in the ensemble.
Instead, it reduces the likelihood of obtaining poor performance.
This phenomenon arises because diverse models tend to make different errors for each sample, but generally concur on correct classifications~\cite{polikar2012ensemble}. 
Illustrated in \cref{fig:qualitative-model-fusion} are predictions made by multiple models and the \mvhc fusion strategy on various \gls*{lp} images.
It is remarkable that model fusion can produce accurate predictions even in cases where most models exhibit prediction errors.
To clarify, with the \mvhc approach, this occurs when each incorrect sequence is predicted fewer times than the correct one, or in the case of a tie, the correct sequence is predicted with higher~confidence.

Shifting our attention back to \cref{tab:ranking-models}, we note that the majority vote-based strategies yielded comparable results, with the sequence-level approach~(MV) performing marginally better for a given number of 
combined models.
Our analysis indicates that this difference arises in cases where a model predicts one character more or one character less, impacting the majority vote by character position~(MVCP) approach relatively more.
Conversely, selecting the prediction with the highest confidence~(HC) consistently led to inferior results.
This can be attributed to the general tendency of all models to make incorrect predictions also with high confidence, as demonstrated in \cref{fig:qualitative-model-fusion}.

A growing number of authors~\cite{zeni2020weakly,laroca2022first,laroca2022cross,wang2022rethinking} have stressed the importance of also evaluating \gls*{lpr} models in a cross-dataset fashion, as it more accurately simulates real-world scenarios where new cameras are regularly being deployed in new locations without existing systems being retrained as often.
Taking this into account, we present in \cref{tab:results-cross} the results obtained on four distinct datasets, none of which were used during the training of the models\footnote{To train the models, we excluded the few images from the \chineselp dataset that are also found in \clpd (both datasets include internet-sourced~images~\cite{laroca2023do}).}.
These particular datasets are commonly employed for this purpose~\cite{zou2020robust,laroca2021efficient,fan2022improving,silva2022flexible,ke2023ultra}.

\begin{table}[!htb]
\centering
\renewcommand{\arraystretch}{1.05}
\setlength{\tabcolsep}{5pt}
\caption{
Comparison of the results achieved in cross-dataset setups by \numbaselines models individually and through five different fusion strategies. 
The models are listed alphabetically, with the highest recognition rates attained for each dataset highlighted in~bold.
}
\label{tab:results-cross}

\vspace{-2mm}

\resizebox{0.775\linewidth}{!}{%
\begin{tabular}{@{}lccccc@{}}
\toprule
\diagbox[trim=l,innerrightsep=15pt]
{Approach}{\multicolumn{1}{c}{\begin{tabular}[c]{@{}c@{}}Dataset\\\# \glspl*{lp}\phantom{i}\end{tabular}}}   & \multicolumn{1}{c}{\begin{tabular}[c]{@{}c@{}}\openalpreu\\\# $108$\phantom{i}\end{tabular}} & \multicolumn{1}{c}{\begin{tabular}[c]{@{}c@{}}\pku\\\# $2{,}253$\phantom{i}\end{tabular}} & \multicolumn{1}{c}{\begin{tabular}[c]{@{}c@{}}\cdhard\\\# $104$\end{tabular}} & \multicolumn{1}{c}{\begin{tabular}[c]{@{}c@{}}\clpd\\\# $1{,}200$\phantom{i}\end{tabular}} & Average \\ \midrule
\crnet~\cite{silva2020realtime}        & $96.3$\% & $99.1$\%   & \phantom{-}$58.7$\%\phantom{-}     & \phantom{--}$94.2$\%\phantom{--}   & \phantom{--}$87.1$\%\phantom{--} \\
\crnn~\cite{shi2017endtoend}             & $93.5$\% & $98.2$\%   & $31.7$\%     & $89.0$\%   & $78.1$\% \\
\fastocr~\cite{laroca2021towards}         & $\textbf{97.2}$\textbf{\%} & $\textbf{99.2}$\textbf{\%}   & $\textbf{59.6}$\textbf{\%}     & $\textbf{94.4}$\textbf{\%}   & $\textbf{87.6}$\textbf{\%} \\
\grcnn~\cite{wang2017deep}       & $87.0$\%   & $98.6$\% & $38.5$\%     & $87.7$\%   & $77.9$\% \\
\holistic~\cite{spanhel2017holistic}          & $89.8$\% & $98.6$\%   & $11.5$\%     & $90.2$\%   & $72.5$\% \\
\multitaskgabriel~\cite{goncalves2019multitask}            & $85.2$\% & $97.4$\%   & $10.6$\%     & $86.8$\%   & $70.0$\% \\
\rtwoam~\cite{lee2016recursive}          & $88.9$\% & $97.1$\%   & $20.2$\%     & $88.2$\%   & $73.6$\% \\
\rare~\cite{shi2016robust}            & $94.4$\% & $98.3$\%  & $37.5$\%     & $92.4$\%   & $80.7$\% \\
\rosetta~\cite{borisyuk2018rosetta}          & $90.7$\% & $97.2$\%  & $14.4$\%     & $86.9$\%   & $72.3$\% \\
\starnet~\cite{liu2016starnet}       & $\textbf{97.2}$\textbf{\%} & $99.1$\%   & $48.1$\%     & $93.3$\%   & $84.4$\%  \\
\trba~\cite{baek2019what}             & $93.5$\% & $98.5$\%   & $35.6$\%     & $90.9$\%   & $79.6$\%  \\
\vitstrbase~\cite{atienza2021vitstr}       & $89.8$\% & $98.4$\%   & $22.1$\%     & $93.1$\%   & \phantom{i}$75.9$\% \\[3pt] \cdashline{1-6} \\[-9pt] \cdashline{1-6} \\[-6pt]
Fusion HC \topcross{(\textit{top 6})}          & $\accAvgHcbmCrossOpenalprEU$\%   & $\accAvgHcbmCrossPku$\% & $\accAvgHcbmCrossCdhard$\%     & $\accAvgHcbmCrossClpd$\%   & $\accAvgHcbmCrossCross$\% \\
Fusion MV--BM \topcross{(\textit{top 8})}          & $\textbf{\accAvgMvbmCrossOpenalprEU}$\textbf{\%}   & $\textbf{\accAvgMvbmCrossPku}$\textbf{\%} & $\textbf{\accAvgMvbmCrossCdhard}$\textbf{\%}     & $\textbf{\accAvgMvbmCrossClpd}$\textbf{\%}   & $\textbf{\accAvgMvbmCrossCross}$\textbf{\%} \\
Fusion MV--HC \topcross{(\textit{top 8})}       & $\textbf{\accAvgMvhcCrossOpenalprEU}$\textbf{\%}   & $\textbf{\accAvgMvhcCrossPku}$\textbf{\%} & $\textbf{\accAvgMvhcCrossCdhard}$\textbf{\%}     & $\accAvgMvhcCrossClpd$\%   & $\accAvgMvhcCrossCross$\% \\
Fusion MVCP--BM \topcross{(\textit{top 9})}       & $\accAvgMvcpBmCrossOpenalprEU$\%   & $\textbf{\accAvgMvcpBmCrossPku}$\textbf{\%} & $\accAvgMvcpBmCrossCdhard$\%     & $\accAvgMvcpBmCrossClpd$\%   & $\accAvgMvcpBmCrossCross$\% \\
Fusion MVCP--HC \topcross{(\textit{top 9})}       & $\accAvgMvcpHcCrossOpenalprEU$\%   & $\textbf{\accAvgMvcpHcCrossPku}$\textbf{\%} & $\accAvgMvcpHcCrossCdhard$\%     & $\accAvgMvcpHcCrossClpd$\%   & $\accAvgMvcpHcCrossCross$\% \\
\bottomrule
\end{tabular}%
}
\end{table}

These experiments provide further support for the findings presented earlier in this section.
Specifically, both strategies that rely on a majority vote at the sequence level (\mvbm and \mvhc) outperformed the others significantly;
the most notable performance gap was observed in the \cdhard dataset, known for its challenges due to the predominance of heavily tilted \glspl*{lp} and the wide variety of \gls*{lp} layouts (as shown in \cref{fig:samples-public-datasets}).
Interestingly, in this cross-dataset scenario, the \mvbm strategy exhibited slightly superior performance compared to \mvhc.
Surprisingly, the \hc approach failed to yield any improvements in results on any dataset, indicating that the models made errors with high confidence even on \gls*{lp} images extracted from datasets that were not part of their~training.

While our primary focus lies on investigating the improvements in recognition rates achieved through model fusion, it is also pertinent to examine its impact on runtime.
Naturally, certain applications might favor combining fewer models to attain a moderate improvement in recognition while minimizing the increase in the system's running time.
With this in mind, \cref{tab:results-speed} presents the number of \gls*{fps} processed by each model independently and when incorporated into the ensemble.
In addition to combining the models based on their average recognition rate across the datasets, as done in the rest of this section, we also explore combining them based on their processing~speed.

\begin{table}[!htb]
\centering
\caption{
The number of \gls*{fps} processed by each model independently and when incorporated into the ensembles.
On the left, the models are ranked based on their results across the datasets, while on the right they are ranked according to their speed.
The reported time, measured in milliseconds per image, represents the average of $5$~runs.
}
\label{tab:results-speed}

\vspace{-2.5mm}

\resizebox{0.99\linewidth}{!}{
\resizebox{!}{18.25ex}{
\begin{tabular}{@{}lcccccccccc@{}}
\toprule
\multicolumn{1}{c}{\multirow{2}{*}[-1.8pt]{\begin{tabular}[c]{@{}c@{}}Models\\ (ranked by \textbf{accuracy})\end{tabular}}} & \multirow{2}{*}[-2.55pt]{\phantom{i}\mvhc\phantom{i}} & \multicolumn{3}{c}{Individual} & & & & \multicolumn{3}{c}{\phantom{I}Fusion} \\ \cmidrule(l{2pt}r{2pt}){3-5} \cmidrule(l{2pt}r{0pt}){9-11} 
             & & Time & & FPS   & & & & Time & & FPS   \\ \midrule
Top $1$ (\vitstrbase) & $92.4$\% & $\phantom{0}7.3$      & & $137$ & & & & $\phantom{0}7.3$      & & $137$ \\
Top $2$ (+ \starnet) & $94.1$\%     & $\phantom{0}7.1$       & & $141$ & & & & $14.4$     & & $\phantom{0}70$  \\
Top $3$ (+ \trba) & $94.9$\%         & $16.9$      & & $\phantom{0}59$  & & & & $31.3$     & & $\phantom{0}32$  \\
Top $4$ (+ \crnet) & $96.3$\%       & $\phantom{0}5.3$       & & $189$ & & & & $36.6$     & & $\phantom{0}27$  \\
Top $5$ (+ \rare) & $96.6$\%         & $13.0$      & & $\phantom{0}77$  & & & & $49.6$     & & $\phantom{0}20$  \\
Top $6$ (+ \fastocr) & $97.0$\%     & $\phantom{0}3.0$       & & $330$ & & & & $52.6$     & & $\phantom{0}19$  \\
Top $7$ (+ \rosetta) & $97.2$\%      & $\phantom{0}4.6$        & & $219$ & & & & $57.2$     & & $\phantom{0}18$  \\
Top $8$ (+ \holistic) & $97.6$\% & $\phantom{0}2.5$       & & $399$ & & & & $59.7$     & & $\phantom{0}17$  \\
Top $9$ (+ \grcnn) & $97.5$\%        & $\phantom{0}8.5$       & & $117$ & & & & $68.2$     & & $\phantom{0}15$  \\
Top $10$ (+ \rtwoam) & $97.2$\%         & $15.9$      & & $\phantom{0}63$  & & & & $84.2$     & & $\phantom{0}12$  \\
Top $11$ (+ \crnn) & $97.0$\%         & $\phantom{0}2.9$       & & $343$ & & & & $87.1$     & & $\phantom{0}11$  \\
Top $12$ (+ \multitaskgabriel) & $97.0$\%  & $\phantom{0}2.3$       & & $427$ & & & & $89.4$     & & $\phantom{0}11$  \\ \bottomrule
\end{tabular}%
}
\hspace{0.19mm} \vline \hspace{0.35mm}
\resizebox{!}{18.25ex}{
\begin{tabular}{@{}lcccccccccc@{}}
\toprule
\multicolumn{1}{c}{\multirow{2}{*}[-1.8pt]{\begin{tabular}[c]{@{}c@{}}Models\\ (ranked by \textbf{speed})\end{tabular}}} & \multirow{2}{*}[-2.55pt]{\phantom{i}\mvhc\phantom{i}} & \multicolumn{3}{c}{Individual} & & & & \multicolumn{3}{c}{\phantom{I}Fusion} \\ \cmidrule(l{2pt}r{2pt}){3-5} \cmidrule(l{2pt}r{0pt}){9-11}
             & & Time & & FPS   & & & & Time & & FPS   \\ \midrule
Top $1$ (\multitaskgabriel) & $85.9$\%  & $\phantom{0}2.3$      & & $427$ & & & & $\phantom{0}2.3$       & & $427$ \\
Top $2$ (+ \holistic) & $90.2$\%     & $\phantom{0}2.5$       & & $399$ & & & & $\phantom{0}4.9$     & & $206$  \\
Top $3$ (+ \crnn) & $91.1$\%         & $\phantom{0}2.9$      & & $343$  & & & & $\phantom{0}7.8$     & & $129$  \\
Top $4$ (+ \fastocr) & $95.4$\%       & $\phantom{0}3.0$       & & $330$ & & & & $10.8$     & & $\phantom{0}93$  \\
Top $5$ (+ \rosetta) & $96.0$\%         & $\phantom{0}4.6$      & & $219$  & & & & $15.4$     & & $\phantom{0}65$  \\
Top $6$ (+ \crnet) & $96.6$\%     & $\phantom{0}5.3$       & & $189$ & & & & $20.7$     & & $\phantom{0}48$  \\
Top $7$ (+ \starnet) & $96.9$\%      & $\phantom{0}7.1$        & & $141$ & & & & $27.8$     & & $\phantom{0}36$  \\
Top $8$ (+ \vitstrbase) & $96.9$\% & $\phantom{0}7.3$       & & $137$ & & & & $35.0$     & & $\phantom{0}29$  \\
Top $9$ (+ \grcnn) & $97.1$\%        & $\phantom{0}8.5$       & & $117$ & & & & $43.6$     & & $\phantom{0}23$  \\
Top $10$ (+ \rare) & $97.1$\%         & $13.0$      & & $\phantom{0}77$  & & & & $56.6$     & & $\phantom{0}18$  \\
Top $11$ (+ \rtwoam) & $97.1$\%         & $15.9$       & & $\phantom{0}63$ & & & & $72.5$     & & $\phantom{0}14$  \\
Top $12$ (+ \trba) & $97.1$\%  & $16.9$       & & $\phantom{0}59$ & & & & $89.4$     & & $\phantom{0}11$  \\ \bottomrule
\end{tabular} \,%
}
}
\end{table}

Remarkably, fusing the outputs of the three fastest models results in a lower recognition rate~($91.1$\%) than using the best model alone~(\bestIndividual).
Nevertheless, as more methods are included in the ensemble, the gap reduces considerably.
From this observation, we can infer that if attaining the utmost recognition rate across various scenarios is not imperative, it becomes more advantageous to combine fewer but faster models, as long as they perform satisfactorily individually.
According to \cref{tab:results-speed}, combining $4$--$6$ fast models appears to be the optimal choice for striking a better balance between speed and~accuracy.

\section{Conclusions}
\label{sec:conclusions}

This paper examined the potential improvements in \gls*{lpr} results by fusing the outputs from multiple recognition models. 
Distinguishing itself from prior studies, our research explored a wide range of models and datasets in the experiments.
We combined the outputs of different models through straightforward approaches such as selecting the most confident prediction or through majority vote (both at sequence and character levels), demonstrating the substantial benefits of fusion approaches in both intra- and cross-dataset experimental~setups.

In the traditional intra-dataset setup, where we explored eight datasets, the mean recognition rate experienced a significant boost, rising from~\bestIndividual achieved by the best model individually to~\bestFusion when leveraging model fusion.
Essentially, we demonstrate that fusing multiple models reduces considerably the likelihood of obtaining subpar performance on a particular dataset.
In the more challenging cross-dataset setup, where we explored four datasets, the mean recognition rate increased from \bestIndividualCross to rates surpassing \fusionExceeding.
Notably, the optimal fusion approach in both setups was via a majority vote at the sequence~level.

We also conducted an evaluation to analyze the speed/accuracy trade-off in the final approach by varying the number of models included in the ensemble.
For this assessment, we ranked the models in two distinct ways: one based on their recognition results and the other based on their efficiency.
The findings led us to conclude that for applications where the recognition task can tolerate some additional time, though not excessively, an effective strategy is to combine $4$--$6$ fast models.
Employing this approach significantly enhances the recognition results while maintaining the system's efficiency at an acceptable~level.

\iffinal
\section*{\uppercase{Acknowledgments}}

\iffinal
    We thank the support of NVIDIA Corporation with the donation of the Quadro RTX $8000$ GPU used for this~research.
\else
    The acknowledgments are hidden for review.
\fi
\else
\fi

\hypersetup{urlcolor=black}
    
\bibliographystyle{splncs04}
\bibliography{bibtex-short}

\end{document}